\documentclass[10pt,twocolumn,letterpaper]{article}

\usepackage[pagenumbers]{cvpr} 

\usepackage{graphicx}
\usepackage{amsmath}
\usepackage{amssymb}
\usepackage{booktabs}
\usepackage{tabularx}
\usepackage{multirow}
\newcolumntype{Y}{>{\centering\arraybackslash}X}
\usepackage[svgnames]{xcolor}

\usepackage{pifont}
\usepackage[bottom]{footmisc}
\usepackage{algorithm}
\usepackage{algpseudocode}
\usepackage{subcaption}
\usepackage[accsupp]{axessibility}  

\def \path {\mathit{path}}

\definecolor{cvprblue}{rgb}{0.21,0.49,0.74}
\usepackage[pagebackref,breaklinks,colorlinks,citecolor=cvprblue]{hyperref}

\renewcommand{\thefootnote}{\fnsymbol{footnote}}

\usepackage[capitalize]{cleveref}
\crefname{section}{Sec.}{Secs.}
\Crefname{section}{Section}{Sections}
\Crefname{table}{Table}{Tables}
\crefname{table}{Tab.}{Tabs.}

\def\paperID{11488} 
\def\confName{CVPR}
\def\confYear{2024}

\title{OV9D: Open-Vocabulary Category-Level 9D Object Pose and Size Estimation}

\newcommand\blfootnote[1]{%
  \begingroup
  \renewcommand\thefootnote{}\footnote{#1}%
  \addtocounter{footnote}{-1}%
  \endgroup
}

\author{Junhao Cai$^{1}$$^\ast$ \quad Yisheng He$^{2}$$^\ast$ \quad Weihao Yuan$^{2}$$^\dagger$ \quad Siyu Zhu$^{3}$ \quad Zilong Dong$^{2}$ \\ Liefeng Bo$^{2}$ \quad Qifeng Chen$^{1}$\\ 
    ${^1}$Hong Kong University of Science and Technology \quad
    ${^2}$Alibaba Group \quad
    ${^3}$Fudan University 
}

\begin{document}
\maketitle

\blfootnote{$^\ast$Equal contributions. $^\dagger$Corresponding author.}%

\begin{abstract}

This paper studies a new open-set problem, the open-vocabulary category-level object pose and size estimation. Given human text descriptions of arbitrary novel object categories, the robot agent seeks to predict the position, orientation, and size of the target object in the observed scene image. To enable such generalizability, we first introduce OO3D-9D, a large-scale photorealistic dataset for this task. Derived from OmniObject3D, OO3D-9D is the largest and most diverse dataset in the field of category-level object pose and size estimation. It includes additional annotations for the symmetry axis of each category, which help resolve symmetric ambiguity.
Apart from the large-scale dataset, we find another key to enabling such generalizability is leveraging the strong prior knowledge in pre-trained visual-language foundation models. 
We then propose a framework built on pre-trained DinoV2 and text-to-image stable diffusion models to infer the normalized object coordinate space (NOCS) maps of the target instances. This framework fully leverages the visual semantic prior from DinoV2 and the aligned visual and language knowledge within the text-to-image diffusion model, which enables generalization to various text descriptions of novel categories. 
Comprehensive quantitative and qualitative experiments demonstrate that the proposed open-vocabulary method, trained on our large-scale synthesized data, significantly outperforms the baseline and can effectively generalize to real-world images of unseen categories.
The project page is at \url{https://ov9d.github.io}.

\end{abstract}

Vision-based object pose estimation (OPE), which aims to estimate the object's position and orientation from the observed visual information, is a fundamental problem in computer vision and robotics.
This task is widely used in numerous applications such as robotic manipulation~\cite{wang2023object}, augmented reality~\cite{marchand2015pose}, surface reconstruction~\cite{wen2023bundlesdf}, etc.

From the perspective of working scope, the existing work of object pose estimation can be divided into three categories: 1) instance-level~\cite{xiang2018posecnn,wang2019densefusion,he2020pvn3d,he2021ffb6d}, 2) generalized instance-level~\cite{liu2022gen6d}, and 3) category-level~\cite{wang2019normalized,ze2022category}. 
Instance-level OPE deals with a limited number of objects that are fully available in both the training and testing sets. 
Generalized instance-level OPE methods can handle objects that are unseen during the training stage, but these methods usually require complete 3D object models~\cite{liu2022gen6d} or large amounts of templates with ground-truth poses~\cite{liu2022gen6d,shugurov2022osop} during the inference stage, which inevitably limits their application. 
Category-level OPE aims at estimating object poses from the category-canonicalized object frame to the camera frame, allowing the model to perform estimation on novel object instances within the same category without the requirement of object models. However, such methods still suffer from similar problems as instance-level OPE, i.e., they cannot generalize to objects with novel categories.

\begin{figure}[t]
\centering
\includegraphics[width=\linewidth]{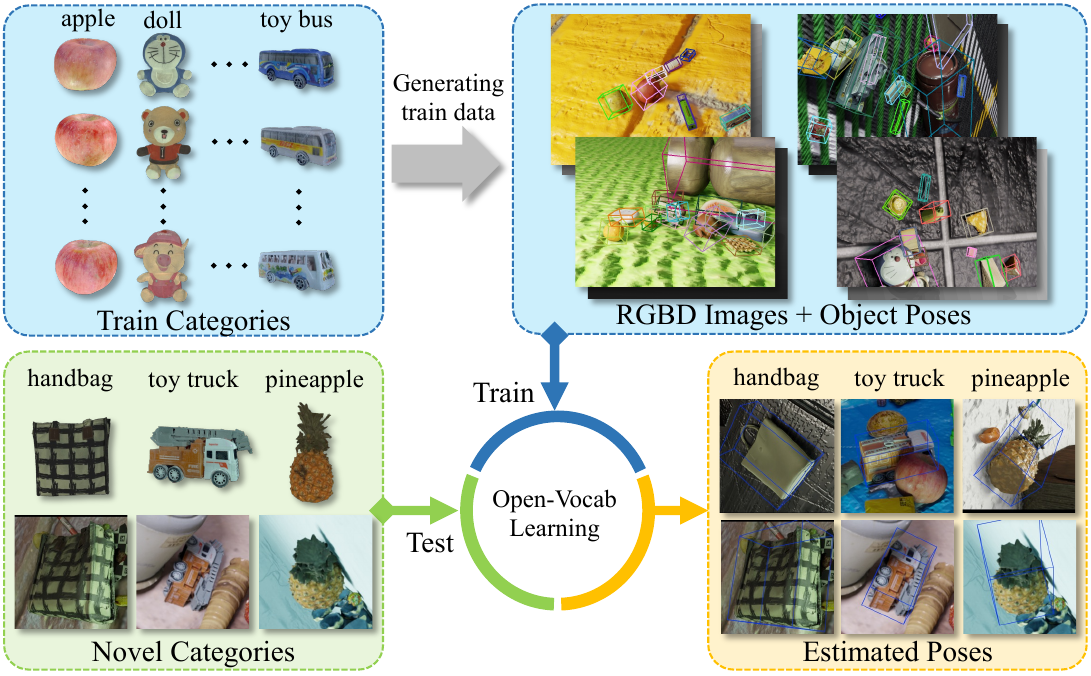}
\vspace{-5mm}
\caption{The open-vocabulary learning of category-level pose and size estimation is trained on a large dataset with diverse categories, such that it could be generalized to novel categories given text prompts of an unseen target object in novel scene images.}
\label{fig:teaser}
\vspace{-5mm}
\end{figure}

While the existing methods of OPE are still confined to limited generalization scope, the rapid development of open-vocabulary learning~\cite{wu2024towards} has demonstrated that open-vocabulary approaches~\cite{du2022learning,liang2023open,xu2023open,zou2024segment,Zhao_2023_ICCV} can effectively eliminate the gap between close-set and open-set scenarios by making use of the feature alignment learned by the visual language models.
These methods have achieved decent performance on many vision tasks such as detection~\cite{du2022learning} and segmentation~\cite{liang2023open,xu2023open,zou2024segment}, even on novel categories in the wild. 
Moreover, many vision foundation models trained on Internet-scale data have also shown remarkable zero-shot performance on many tasks~\cite{xu2023open,Zhao_2023_ICCV,zhang2024tale}. 
However, the task of open-vocabulary pose and size estimation has not been touched so far.

In this paper, we take one step forward in this direction by introducing a new challenging problem: open-vocabulary category-level object pose estimation, which enables estimating poses and size of novel categories in a novel RGB-D scene image with only text inputs. 
To enable such generalization capability, our key insight is that we can take advantage of the prior alignment knowledge in visual-language foundation models pre-trained from large-scale language and image datasets.
This alignment can guide the training for consistent pose estimation using identical text descriptions for same-category instances and enable knowledge to transfer to new categories. Specifically, we design an open-vocabulary framework based on CLIP (Contrastive Language-Image Pretraining)~\cite{radford2021learning}, text-to-image stable diffusion model~\cite{rombach2022high}, and DinoV2~\cite{oquab2023dinov2} to estimate the normalized object coordinate space (NOCS) maps~\cite{wang2019normalized} from the monocular RGB images. Concretely, text features from the CLIP model and latent visual features from VQVAE~\cite{esser2021taming} are fused together in the diffusion UNet to generate diffusion features. The RGB images are also fed into the DinoV2 module to extract additional discriminative features. These two types of features are jointly leveraged to estimate the NOCS maps for the target objects.

For the training of this framework, current datasets~\cite{sundermeyer2023bop,wang2019normalized,ze2022category}, however, contain either limited object instances or few categories, which are not enough for open-vocabulary learning.
To resolve this, we create a large-scale photo-realistic dataset, namely OO3D-9D, which is derived from OmniObject3D (OO3D)~\cite{wu2023omniobject3d}.
This dataset comprises 5,371 objects spanning 216 categories, encompassing both single-object and multi-object scenarios. 
Each single-object scenario is composed of 1,000 RGB-D images with ground truth object poses.
Each multi-object scenario is composed of 5 captures of 5-20 objects, and there are 100K various multi-object scenes in total.

In the experiments, we evaluate our method using both synthesized data and real data from Co3D~\cite{reizenstein2021common}. 
The results demonstrate that our approach, trained with the generated large-scale dataset, achieves notable performance on novel object instances as well as objects belonging to previously unseen categories. 

Our main contributions are then summarized as follows:
\begin{itemize}
    \item We introduce a new challenging problem, the open-vocabulary category-level object pose and size estimation, and establish a benchmark to study it. 
    
    \item We introduce a large-scale photo-realistic dataset for model training in this problem. 
    This synthesized dataset is the largest and most diverse dataset for category-level object pose and size estimation. 
    To generate this dataset, we add annotations of the objects' symmetry axes in OO3D, making it suitable for addressing our problem. 
    
    \item We propose an open-vocabulary framework to tackle this problem, where we fully leverage the visual semantic prior from the pre-trained Dino and the aligned visual and text knowledge within the text-to-image diffusion model. For the first time, we reveal that these priors can boost the network generalizability to real-world images of unseen categories and enable open-vocabulary pose and size estimation given free-form descriptions of the target object.
\end{itemize}
\section{Related Work}

\subsection{Generalized Object Pose Estimation}
The task of generalized instance-level object pose estimation focuses on estimating 6-degree-of-freedom (6-DoF) poses for novel objects that are not available during training or finetuning. 
Traditional methods follow the pipeline of template rendering, feature extraction, and template matching to find the matched template and point correspondences~\cite{hinterstoisser2013model,konishi2016fast,konishi2019real}. 
To extract more robust features, recent methods use deep neural networks to perform object detection and correspondence matching between the reference and query images~\cite{shugurov2022osop,he2022fs6d,liu2022gen6d,zhao2022fusing}. 
Although these methods provide a more general framework to estimate poses for unseen objects, they inevitably require object models or templates with ground-truth 6-DoF poses, which might be impractical in real-world applications. 

\subsection{Category-Level Object Pose Estimation}
Category-level object poses estimation~\cite{bruns2023rgb} aims to estimate the scale, position, and orientation of the objects in the same category that shares the same reference frame, which is a more general task compared with instance-level object pose estimation~\cite{xiang2018posecnn}. 
Early work investigates this task by jointly detecting the object and estimating the corresponding normalized coordinates in normalized object coordinate space (NOCS)~\cite{wang2019normalized}. 
Follow-up works leverage the geometric information from the observed point cloud or object shape prior to estimate both the canonical object coordinates and object poses~\cite{chen2020learning,heTSS9D,irshad2022centersnap,tian2020shape,wang2021category,chen2021sgpa,lin2022category,li2023generative}. 
Other methods solve this problem in an analysis-by-synthesis manner by comparing the rendered and observed images~\cite{chen2020category,lee2021category,deng2022icaps,bruns2022sdfest}. While these methods can only deal with seen object categories during training, our open-vocabulary framework can generalize to unseen categories given the text descriptions.

\begin{figure}[t]
\centering
\includegraphics[width=0.8\linewidth]{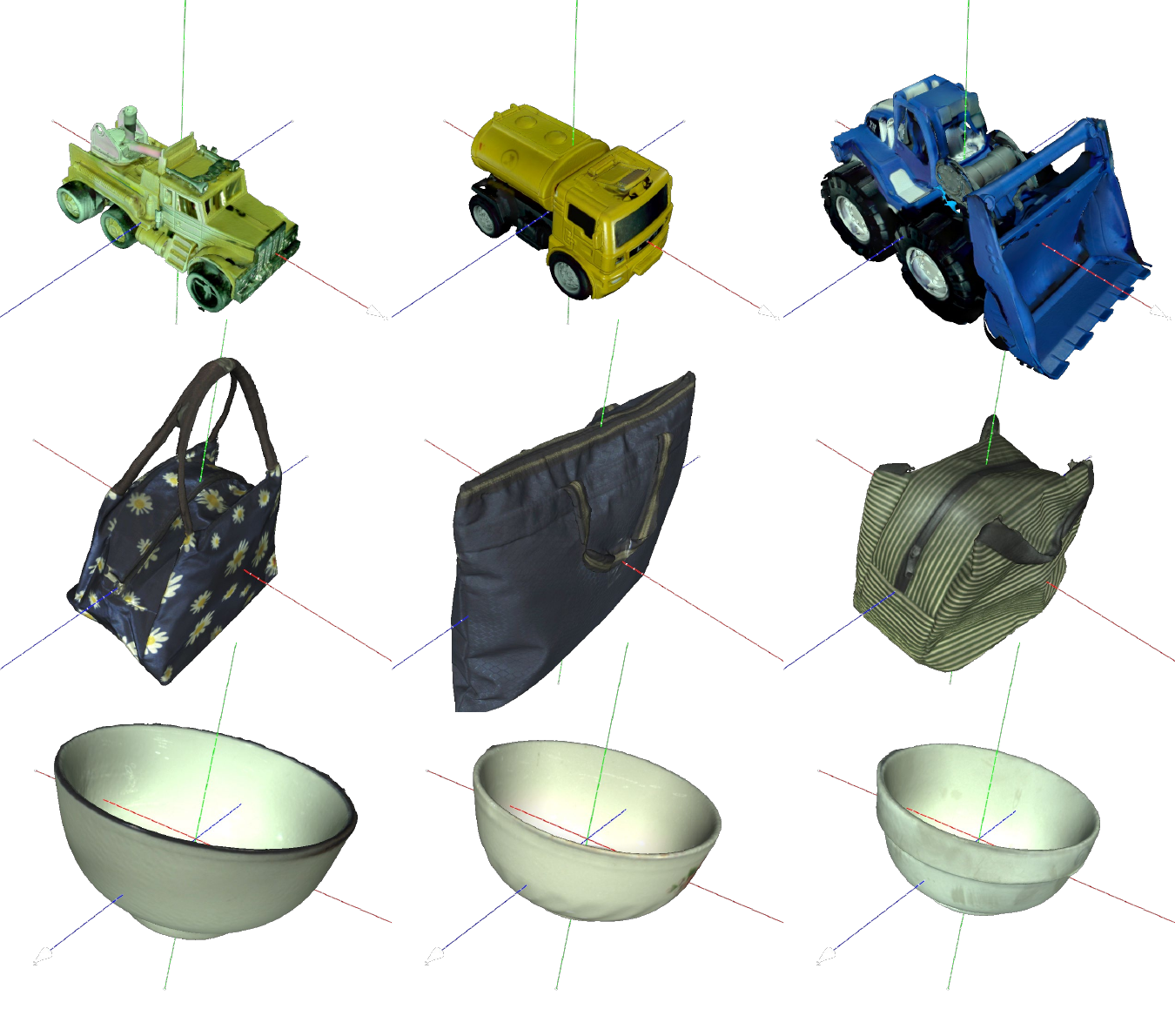}
\vspace{-5mm}
\caption{Visualization of aligned objects: Row 1 features non-symmetric toy trucks with their heads aligned to the X-axis. Row 2 presents handbags containing discrete symmetric axes, with their openings aligned to the Y-axis. Row 3 shows bowls that possess continuous symmetric axes.}
\label{fig:objects_with_aligned_body_frames}
\vspace{-5mm}
\end{figure}

\subsection{Datasets}
Although the category-level methods perform well on novel instances within the same category, they cannot be generalized to objects of a novel category. One main challenge is the lack of training data. 
The most widely used benchmark is REAL275 and CAMERA25 proposed by~\cite{wang2019normalized}, which contains only 6 categories and 18 object instances. Wild6D~\cite{ze2022category} scales up the number of object instances to 1722, but the categories' diversity is still limited to 5 categories~\cite{ze2022category}. On the other hand, large amounts of 3D object datasets have been proposed for various tasks including shape reconstruction and generation~\cite{shapenet2015,reizenstein2021common,downs2022google,deitke2023objaverse,wu2023omniobject3d}. OO3D~\cite{wu2023omniobject3d} introduced an aligned real-scanned 3D database including 6000 objects in 190 categories, which is diverse enough to cover the commonly seen objects in our daily life. However, this dataset only contains 3D models and the object symmetry information is lost. In this work, we add annotations of the objects' symmetry axes and render a large-scale dataset for open-vocabulary pose and size estimation. Our dataset is so far the largest and most diverse dataset for category-level object pose and size estimation. 

\subsection{Open-Vocabulary Learning and Vision Foundation Models}
Recently, visual language models trained with Internet-scaled data have shown remarkable zero-shot performance on various vision tasks~\cite{radford2021learning,wu2024towards}. These methods, including open-vocabulary object detection~\cite{du2022learning}, segmentation~\cite{liang2023open,xu2023open,zou2024segment}, and depth estimation~\cite{Zhao_2023_ICCV}, make use of the alignment learned in the visual language models to reduce the gap between in-distribution and out-of-distribution scenarios. Meanwhile, many research results have demonstrated that visual models including text-to-image diffusion models~\cite{rombach2022high,saharia2022photorealistic} or ViT~\cite{caron2021emerging,oquab2023dinov2} trained with billions of data~\cite{schuhmann2022laion} can serve as generalized feature descriptors, which remarkably improve the performance of generalized tasks including open-vocabulary learning~\cite{xu2023open,Zhao_2023_ICCV} and correspondence matching~\cite{zhang2024tale,tang2024emergent}. Differently, we exploit features extracted from both stable diffusion~\cite{rombach2022high} and DINOv2~\cite{oquab2023dinov2} to infer the objects' normalized coordinates and estimate the pose and size parameters. To the best of our knowledge, we are the first to show that these priors can enable the open-vocabulary pose and size estimation given free-form descriptions of the target object.

\begin{figure}[]
\centering
\includegraphics[width=\linewidth]{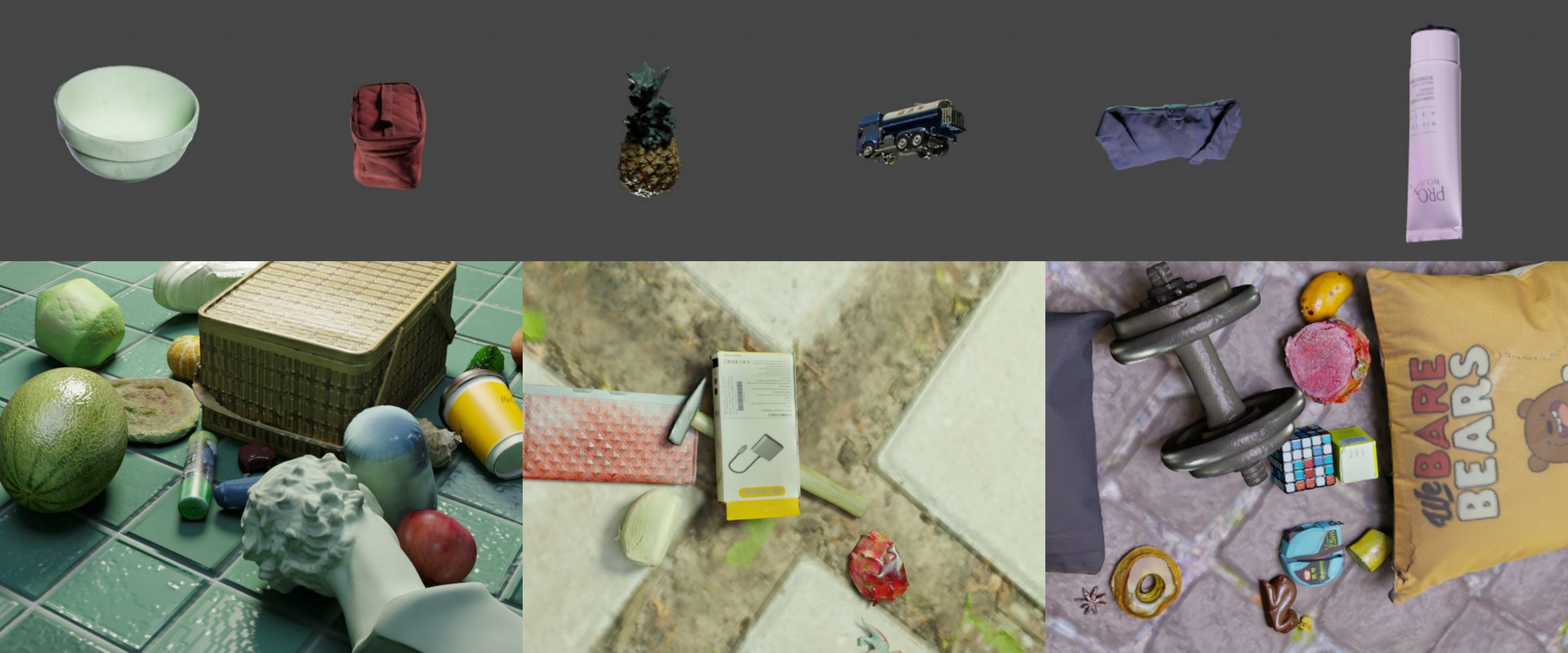}
\vspace{-7mm}
\caption{Example images in OO3D-9D dataset. Single-object scenes as CO3D are displayed in the first row while challenging multi-object scenes are displayed in the second row.}
\label{fig:oo3d_9d_example}
\vspace{-3mm}
\end{figure}

\begin{figure*}[h]
\centering
\includegraphics[width=0.98\linewidth]{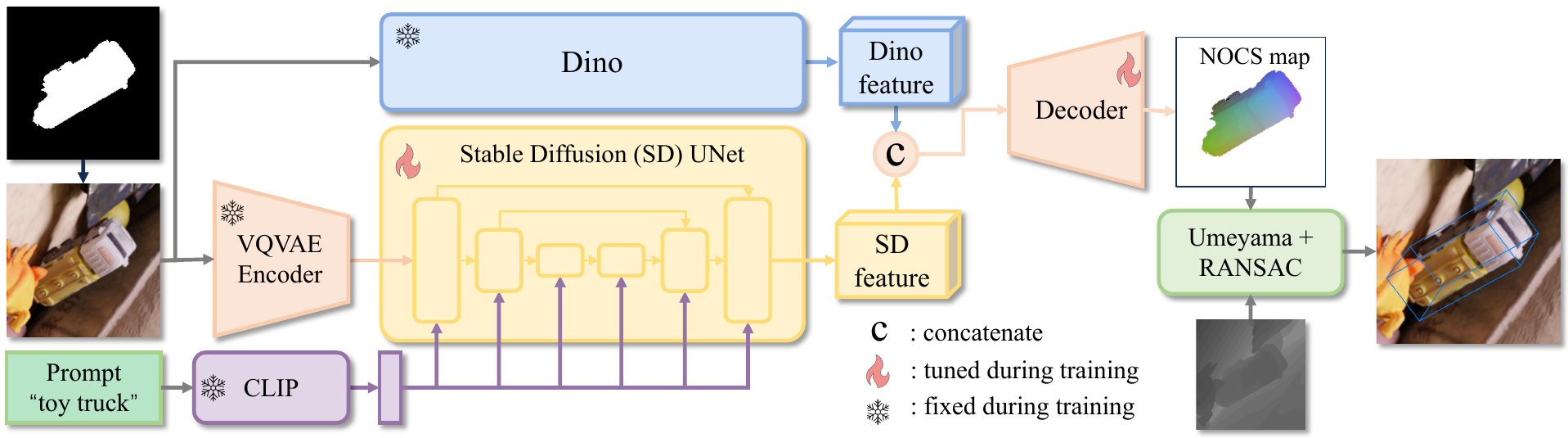}
\vspace{-3mm}
\caption{Overall framework. Text features are acquired from the prompt through the CLIP model and fed to the SD UNet. By combining these text features with latent visual features from VQVAE, SD feature maps are generated. Simultaneously, the DinoV2 module processes the masked RGB image to obtain Dino features. Both features are then combined in the decoder to estimate the NOCS map of the target object. During the inference stage, the depth map is utilized to establish correspondence between NOCS and the camera frame. Finally, the object's size and pose are computed using a pose-fitting algorithm.}
\label{fig:framework}
\vspace{-3mm}
\end{figure*}

\section{Methodology}
\label{Section:Approach}


\subsection{Problem Statement}
\label{Subsec:Prob:Statement}

In this paper, we consider the task of open-vocabulary category-level object pose estimation as evaluating the scale, position, and orientation of the objects in novel categories that are not present in the training stage. 
For the objects within the same category, the model should be able to recognize them and assign them with implicitly consistent reference frames. In other words, the body frames attached to these objects are aligned for scale, position, and orientation under the cognition of the model. Figure~\ref{fig:objects_with_aligned_body_frames} gives an example of such alignment. 

Formally, suppose an object set is divided into $m+n$ subsets as
\begin{equation}
    \mathcal{O}=\mathcal{O}_M \bigcup \mathcal{O}_N
\end{equation}
where $\mathcal{O}_M=\mathcal{O}_1 \bigcup \mathcal{O}_2 \bigcup ... \bigcup \mathcal{O}_m$, and $\mathcal{O}_N=\mathcal{O}_{m+1}\bigcup ... \bigcup \mathcal{O}_{m+n}$. 
Each subset $\mathcal{O}_i$ has a text description $d_i$ characterizing the category of the object set and a reference frame ${}^{\mathcal{O}_i}T$ shared by all the instances in the set. For a learning-based method, we further assume that the first $m$ subsets are the training set, and the remaining $n$ subsets are unknown for the model. Therefore, the task of open-vocabulary category-level object pose estimation could be formulated as estimating a 9-DoF size and pose ${}^cT_{\mathcal{O}_i} = \{s, {}^cR_{\mathcal{O}_i}, t_c\}$ from the observed quadruple $\{I_{o_{ij}}, D_{o_{ij}}, M_{o_{ij}}, d_i\}$, where $s$ is the 3-D scale vector, ${}^cR_{\mathcal{O}_i}$ and $t_c$ are the rotation and translation components, $o_{ij} \in \{O_i\}$, $\{I_{o_{ij}}, D_{o_{ij}}\}$ is the RGBD pair containing the object, $M_{o_{ij}}$ is the object mask, and $i \in [1, m+n]$. It is worth noting that in the ideal case, for any two objects in the same category $o_{ij},o_{ik} \in \{O_i\}$, the intra-category frame consistency requires ${}^{o_{ik}}T_{o_{ij}} = I$.

\subsection{Datasets}
\label{Subsec:App:Dataset}
To ensure the network generalization capability, we generate a new large-scale 
and diverse pose dataset based on OmniObject3D~\cite{wu2023omniobject3d} and BlenderProc~\cite{denninger2023blenderproc2}, called OO3D-9D. Overall, OO3D-9D consists of 5371 object instances in 216 categories. Each instance contains 1000 RGBD image pairs with ground truth 3D object size, 6-DoF poses, and masks. Among this dataset, we draw 10 categories of 230 instances as the test set of novel categories. We further select 214 instances from the known categories as the test set of novel instances. We further generate a multi-object dataset among these objects. Table~\ref{tab:oo3d_9d} provides an overview of OO3D-9D and comparisons with other datasets. 

\begin{table}[h]
\centering
\small
\scalebox{0.95}{
\begin{tabular}{l|c|c|c|c}
\hline
Datasets & \#instance & \#category & \#image & GT \\ \hline
CAMERA25~\cite{wang2019normalized} & 184 & 6 & 300K & $\checkmark$ \\ \hline
REAL275~\cite{wang2019normalized} & 24 & 6 & 8K & $\checkmark$ \\ \hline
Wild6D-train~\cite{ze2022category} & 1,560 & 5 & 1M & $\times$ \\ \hline
Wild6D-test~\cite{ze2022category} & 162 & 5 & 100K & $\checkmark$ \\ \hline
FS6D~\cite{he2022fs6d} & \textbf{12,490} & 51 & 800K &$\checkmark$ \\ \hline
OO3D-9D & 5,371 & \textbf{216} & \textbf{5M} & $\checkmark$ \\ \hline
\end{tabular}}
\vspace{-2mm}
\caption{Comparison of different datasets. Our OO3D-9D is the largest and most diverse in object categories, which is critical in network generalizability under our open-vocabulary setting. GT denotes ground-truth object pose and size annotation.}
\vspace{-3mm}
\label{tab:oo3d_9d}
\end{table}

\subsubsection{Symmetry Axis Annotation of OO3D}
While OmniObject3D contains large amounts of objects with diverse categories, the symmetric property of the objects is not provided, which is essential information for handling symmetry-shaped objects for the task of object pose estimation. To deal with this problem, we manually annotate symmetry axes for objects based on their shapes. Concretely, we follow the BOP data format~\cite{sundermeyer2023bop} and add transformation matrices for discrete symmetry and rotation axes for continuous symmetry. Overall, the dataset comprises 1129 objects exhibiting discrete symmetry and 2066 objects displaying continuous symmetry. Figure~\ref{fig:objects_with_aligned_body_frames} also visualizes some objects with their symmetry axes. 

It is worth noting that many objects in OmniObject3D, such as apples and broccoli, neither have a clear sense of direction nor exhibit strict symmetry. For these objects, we annotate continuous symmetric axes along recognizable directions and consider them as the principal directions of the objects. Consequently, the model will estimate only the principal directions when dealing with such objects.

\subsubsection{OO3D-9D Generation}
We follow the usage of BlenderProc~\cite{denninger2023blenderproc2} by BOP benchmark~\cite{sundermeyer2023bop} to render data for each object. Specifically, RGBD images with object masks and camera poses are rendered from randomized viewpoints sampled on the spheres whose radius ranges between 3 and 5 times the diagonal size of the object. The principal directions of the cameras always point toward the object center. Such configuration not only ensures a well-distributed pose space but also guarantees a clear view of the objects. 

To construct a more comprehensive benchmark, we follow the pipeline of photorealistic image synthesis used in BOP Challenge~\cite{sundermeyer2023bop} to further generate a multi-object dataset. Concretely, a random number of objects are sampled from the object set and placed on the workspace with random poses. We then randomly select one of the points of interest based on the location of the objects as the viewpoint of the camera. The visual data with ground truth poses is finally rendered with randomized lighting conditions and background textures. To simplify the usage, both datasets are organized in BOP format~\cite{sundermeyer2023bop}. 

\subsection{Approach}
\label{Subsec:App:Aproach}

\subsubsection{Preliminaries}

\noindent\textbf{CLIP}~\cite{radford2021learning}. CLIP (Contrastive Language-Image Pretraining) is a visual language model based on a transformer architecture~\cite{vaswani2017attention} pre-trained using a contrastive learning approach that maximizes the agreement between semantically related images and text while minimizing the agreement between unrelated pairs. 

\noindent\textbf{Diffusion model}~\cite{rombach2022high}. The text-to-image Diffusion model is a generative framework that synthesizes high-quality images from text descriptions. The architecture of the diffusion model compromises three modules: an encoder and a decoder derived from VQVAE~\cite{esser2021taming} to map features between image and latent spaces, and a U-Net that performs a denoising operation on the latent space. 

\noindent\textbf{Dino}~\cite{caron2021emerging,oquab2023dinov2}. Dino and DinoV2 are self-supervised learning methods for visual representation learning. The model is based on Vision Transformers (ViT)~\cite{dosovitskiy2020vit} and employs contrastive learning on large-scale datasets, which is encouraged to learn discriminative and diverse features.  

\subsubsection{Overview} 

The proposed method consists of a neural network $f_{\theta}(I, d)$ that takes as input a masked RGB image $I$ and the text description $d$ of the object and outputs an estimation of the normalized object coordinates map $Y$ of that object in the normalized object coordinate space (NOCS)~\cite{wang2019normalized}. The learning objective is to optimize the prediction of $Y$ such that by leveraging the depth map and camera intrinsic, we can estimate scales, rotation, and translation from the NOCS to the camera frame. 

The overview of the network $f$ is shown in Figure.~\ref{fig:framework}, which consists of five parts: 1) the encoder of VQVAE $f_{{\theta}_v}$ that maps the image features to the latent space, 2) the CLIP model $f_{{\theta}_c}$ that encodes the text description to latent features, 3) the denoising UNet $f_{{\theta}_u}$ in Stable Diffusion that captures semantic relationships between text and visual features, 4) the DinoV2 model $f_{{\theta}_d}$ that extracts discriminative features from the images, and 5) the NOCS decoder $f_{{\theta}_n}$ that aggregates the features together and predicts the NOCS map of the object. 

\subsubsection{Normalized Object Coordinate Estimation From Prior Foundation Model}

The encoder builds upon VPD~\cite{Zhao_2023_ICCV} and consists of 4 parts. Latent visual features $v_v=f_{{\theta}_v}(I)$ and $v_c=f_{{\theta}_c}(d)$ text features are first extracted by the encoder of VQVAE and the CLIP model respectively. These two features are then incorporated together by the cross-attention modules in the denoising UNet and result in semantic features $v_s=f_{{\theta}_u}(v_v, v_c)$ of the target objects. Since the estimation of normalized object coordinates requires the model to recognize every part of the object in a discriminative manner, we further employ DinoV2 on the RGB images to extract additional image features $v_d=f_{{\theta}_d}(I)$. 

With $v_s$ and $v_d$ in place, we concatenate and aggregate them by a convolutional layer. In the implementation of the diffusion UNet, the resolution of the output feature maps is $1/32$ times the size of the original image, we thus use a 5-layer fully convolutional residual network as the decoder network~\cite{he2016deep} where each layer contains a bilinear $2\times$ upsampling module such that it outputs a coordinate map $\hat{Y}=f_{{\theta}_n}(v_s, v_d)$ with the same size and resolution as that of the input image.  

The model is trained by minimizing the smooth L1 loss~\cite{girshick2015fast} between the predicted and ground truth normalized coordinate maps, which is denoted as
\begin{equation}
\label{eqn:smooth_l1_loss}
L=\frac{1}{N} \sum_{i=1}^{H}\sum_{j=1}^{W} M_{ij}S_{ij},
\end{equation}
where $M$ is the object mask, $N=\sum_{i=1}^{H}\sum_{j=1}^{W} M_{ij}$, and $S_{ij}$ is the smooth L1 loss at pixel $ij$ 
\begin{equation}
\label{eqn:pixelwise_smooth_l1_loss}
\begin{split}
S_{ij}= \left \{
\begin{array}{ll}
  0.5(Y_{ij}-\hat{Y}_{ij})^2/\beta, & |Y_{ij}-\hat{Y}_{ij}| < \beta ,\\
  |Y_{ij}-\hat{Y}_{ij}| - 0.5\beta, & otherwise
\end{array},
\right.
\end{split}
\end{equation}
where $\beta$ is the smooth threshold and is set to 0.1 in this work. 

For objects with symmetry, we will first augment the ground truth by transforming the ground truth coordinate map according to the symmetric axes, which form an augmented coordinate map set $\mathcal{S}=\{S\}$. Then the corresponding loss function becomes

\begin{equation}
\label{eqn:smooth_l1_loss_with_symmetry}
L=\inf_{S \in \mathcal{S}}[\frac{1}{N} \sum_{i=1}^{H}\sum_{j=1}^{W} M_{ij}S_{ij}].
\end{equation}

\subsubsection{Pose and Size Estimation}
Given the observed depth map with the camera intrinsic, the partial point cloud of the object in the camera frame can be obtained. With the estimated normalized coordinates, we can further establish the 3D correspondences between the camera frame and the normalized coordinate space. Following the pipeline in~\cite{wang2019normalized}, we can recover the object pose by using the Umeyama algorithm~\cite{umeyama1991least} with RANSAC paradigm~\cite{fischler1981random}. 

\subsubsection{Network Architecture}
We use the stable diffusion model~\cite{rombach2022high} pre-trained on LAION-5b~\cite{schuhmann2022laion} as one of the image feature extractors, where we only use the feature maps extracted from the last block of the denoising UNet. Following VPD~\cite{Zhao_2023_ICCV} and ODISE~\cite{xu2023open}, we add zero noise to the input feature maps by setting the time step to 0. 
For DinoV2~\cite{oquab2023dinov2}, we use the distilled version ViT-L whose output stride is set to 14. We further perform bilinear interpolation on the Dino features to make the output image features consistent with features extracted from the diffusion UNet. As for the CLIP model~\cite{radford2021learning}, we only use the text encoder to extract the text features. 

All these modules are initialized with the corresponding pre-trained models. Only the decoder is initialized with Kaiming Initialization.
To fully exploit the performance of the pre-trained models, we keep CLIP, DinoV2, and the encoder of VQVAE fixed and only tune the parameters of the diffusion UNet and the decoder during training.

\begin{figure*}[h]
\centering
\includegraphics[width=0.9\linewidth]{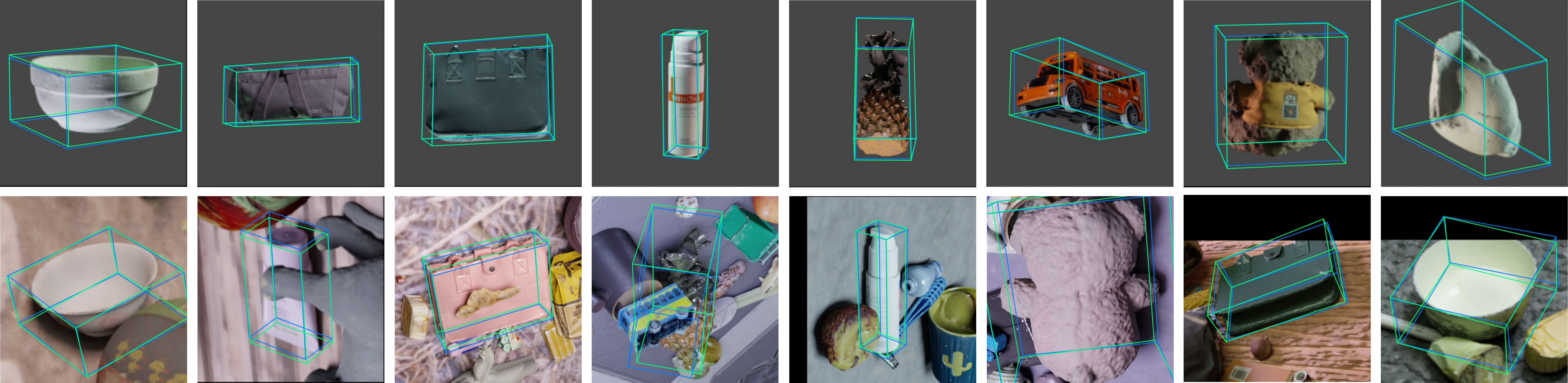}
\vspace{-3mm}
\label{fig:vis_results_oo3d}
\caption{Qualitative results on OO3D-9D. Row 1: single object scene. Row 2: occluded scene. Predicted and ground truth boxes are colored in blue and green, respectively.}
\vspace{-3mm}
\end{figure*}

\begin{table*}[t]
\centering
\small
\scalebox{0.75}{
\begin{tabular}{c|ccc|ccc|ccc|ccc|ccc|ccc|ccc}
\hline
\multirow{2}{*}{} & \multicolumn{3}{c|}{Abs IoU@50} & \multicolumn{3}{c|}{Abs $5^\circ5cm$} & \multicolumn{3}{c|}{Abs $10^\circ5cm$} & \multicolumn{3}{c|}{Abs $10^\circ10cm$} & \multicolumn{3}{c|}{Rel $5^\circ5cm$} & \multicolumn{3}{c|}{Rel $10^\circ5cm$} & \multicolumn{3}{c}{Rel $10^\circ10cm$} \\ \cline{2-22} 
 & \multicolumn{1}{c|}{PCA} & \multicolumn{1}{c|}{IST} & Ours & \multicolumn{1}{c|}{PCA} & \multicolumn{1}{c|}{IST} & Ours & \multicolumn{1}{c|}{PCA} & \multicolumn{1}{c|}{IST} & Ours & \multicolumn{1}{c|}{PCA} & \multicolumn{1}{c|}{IST} & Ours & \multicolumn{1}{c|}{PCA} & \multicolumn{1}{c|}{IST} & Ours & \multicolumn{1}{c|}{PCA} & \multicolumn{1}{c|}{IST} & Ours & \multicolumn{1}{c|}{PCA} & \multicolumn{1}{c|}{IST} & Ours \\ \hline
bowl$\dagger$ & \multicolumn{1}{c|}{19.2} & \multicolumn{1}{c|}{53.7} & \textbf{97.3} & \multicolumn{1}{c|}{1.2} & \multicolumn{1}{c|}{14.7} & \textbf{41.7} & \multicolumn{1}{c|}{1.3} & \multicolumn{1}{c|}{32.5} & \textbf{62.3} & \multicolumn{1}{c|}{1.3} & \multicolumn{1}{c|}{32.5} & \textbf{62.3} & \multicolumn{1}{c|}{2.3} & \multicolumn{1}{c|}{1.0} & \textbf{66.3} & \multicolumn{1}{c|}{4.7} & \multicolumn{1}{c|}{2.7} & \textbf{90.5} & \multicolumn{1}{c|}{9.3} & \multicolumn{1}{c|}{2.7} & \textbf{90.5} \\ \hline
bumbag$\ddagger$ & \multicolumn{1}{c|}{29.6} & \multicolumn{1}{c|}{40.4} & \textbf{91.7} & \multicolumn{1}{c|}{2.4} & \multicolumn{1}{c|}{0.0} & \textbf{10.9} & \multicolumn{1}{c|}{5.2} & \multicolumn{1}{c|}{0.2} & \textbf{41.5} & \multicolumn{1}{c|}{5.8} & \multicolumn{1}{c|}{0.2} & \textbf{41.5} & \multicolumn{1}{c|}{2.4} & \multicolumn{1}{c|}{0.2} & \textbf{14.2} & \multicolumn{1}{c|}{4.0} & \multicolumn{1}{c|}{0.2} & \textbf{44.2} & \multicolumn{1}{c|}{11.8} & \multicolumn{1}{c|}{0.2} & \textbf{44.2} \\ \hline
dumpling$\ddagger$ & \multicolumn{1}{c|}{46.7} & \multicolumn{1}{c|}{64.5} & \textbf{81.2} & \multicolumn{1}{c|}{\textbf{3.7}} & \multicolumn{1}{c|}{0.0} & 1.8 & \multicolumn{1}{c|}{6.6} & \multicolumn{1}{c|}{0.4} & \textbf{11.0} & \multicolumn{1}{c|}{6.6} & \multicolumn{1}{c|}{0.4} & \textbf{11.0} & \multicolumn{1}{c|}{\textbf{6.0}} & \multicolumn{1}{c|}{0.1} & 1.9 & \multicolumn{1}{c|}{\textbf{11.3}} & \multicolumn{1}{c|}{0.1} & 10.4 & \multicolumn{1}{c|}{\textbf{11.3}} & \multicolumn{1}{c|}{0.1} & 10.4 \\ \hline
facial cream$\ddagger$ & \multicolumn{1}{c|}{3.6} & \multicolumn{1}{c|}{37.1} & \textbf{91.3} & \multicolumn{1}{c|}{0.4} & \multicolumn{1}{c|}{7.7} & \textbf{34.0} & \multicolumn{1}{c|}{1.7} & \multicolumn{1}{c|}{17.1} & \textbf{52.7} & \multicolumn{1}{c|}{1.7} & \multicolumn{1}{c|}{17.1} & \textbf{52.7} & \multicolumn{1}{c|}{2.1} & \multicolumn{1}{c|}{0.3} & \textbf{25.3} & \multicolumn{1}{c|}{4.9} & \multicolumn{1}{c|}{2.3} & \textbf{56.4} & \multicolumn{1}{c|}{4.9} & \multicolumn{1}{c|}{2.3} & \textbf{58.4} \\ \hline
handbag$\ddagger$ & \multicolumn{1}{c|}{35.5} & \multicolumn{1}{c|}{52.3} & \textbf{91.8} & \multicolumn{1}{c|}{1.6} & \multicolumn{1}{c|}{0.0} & \textbf{22.0} & \multicolumn{1}{c|}{3.0} & \multicolumn{1}{c|}{0.2} & \textbf{54.7} & \multicolumn{1}{c|}{4.1} & \multicolumn{1}{c|}{0.2} & \textbf{56.0} & \multicolumn{1}{c|}{4.1} & \multicolumn{1}{c|}{0.0} & \textbf{25.3} & \multicolumn{1}{c|}{6.8} & \multicolumn{1}{c|}{0.1} & \textbf{56.4} & \multicolumn{1}{c|}{10.0} & \multicolumn{1}{c|}{0.1} & \textbf{56.4} \\ \hline
litchi$\dagger$ & \multicolumn{1}{c|}{32.4} & \multicolumn{1}{c|}{89.9} & \textbf{94.6} & \multicolumn{1}{c|}{1.7} & \multicolumn{1}{c|}{15.3} & \textbf{7.0} & \multicolumn{1}{c|}{2.8} & \multicolumn{1}{c|}{26.4} & \textbf{18.6} & \multicolumn{1}{c|}{2.8} & \multicolumn{1}{c|}{26.4} & \textbf{18.6} & \multicolumn{1}{c|}{2.4} & \multicolumn{1}{c|}{0.4} & \textbf{18.4} & \multicolumn{1}{c|}{3.7} & \multicolumn{1}{c|}{2.5} & \textbf{40.5} & \multicolumn{1}{c|}{3.7} & \multicolumn{1}{c|}{2.5} & \textbf{40.5} \\ \hline
mouse$\ast$ & \multicolumn{1}{c|}{10.8} & \multicolumn{1}{c|}{28.8} & \textbf{72.9} & \multicolumn{1}{c|}{0.3} & \multicolumn{1}{c|}{0.0} & \textbf{6.3} & \multicolumn{1}{c|}{0.8} & \multicolumn{1}{c|}{0.1} & \textbf{17.7} & \multicolumn{1}{c|}{0.8} & \multicolumn{1}{c|}{0.1} & \textbf{17.7} & \multicolumn{1}{c|}{2.7} & \multicolumn{1}{c|}{0.1} & \textbf{3.9} & \multicolumn{1}{c|}{7.3} & \multicolumn{1}{c|}{0.1} & \textbf{15.0} & \multicolumn{1}{c|}{7.3} & \multicolumn{1}{c|}{0.1} & \textbf{15.0} \\ \hline
pineapple$\dagger$ & \multicolumn{1}{c|}{29.7} & \multicolumn{1}{c|}{41.0} & \textbf{89.4} & \multicolumn{1}{c|}{1.4} & \multicolumn{1}{c|}{6.9} & \textbf{30.8} & \multicolumn{1}{c|}{2.8} & \multicolumn{1}{c|}{14.2} & \textbf{49.5} & \multicolumn{1}{c|}{3.2} & \multicolumn{1}{c|}{15.0} & \textbf{49.5} & \multicolumn{1}{c|}{1.9} & \multicolumn{1}{c|}{0.2} & \textbf{47.3} & \multicolumn{1}{c|}{3.4} & \multicolumn{1}{c|}{0.8} & \textbf{69.0} & \multicolumn{1}{c|}{3.4} & \multicolumn{1}{c|}{0.8} & \textbf{72.3} \\ \hline
teddy bear$\ast$ & \multicolumn{1}{c|}{64.0} & \multicolumn{1}{c|}{61.8} & \textbf{91.6} & \multicolumn{1}{c|}{0.2} & \multicolumn{1}{c|}{0.0} & \textbf{1.0} & \multicolumn{1}{c|}{0.4} & \multicolumn{1}{c|}{0.6} & \textbf{5.0} & \multicolumn{1}{c|}{0.6} & \multicolumn{1}{c|}{0.6} & \textbf{5.2} & \multicolumn{1}{c|}{\textbf{2.0}} & \multicolumn{1}{c|}{0.2} & 1.0 & \multicolumn{1}{c|}{\textbf{3.0}} & \multicolumn{1}{c|}{0.2} & 2.0 & \multicolumn{1}{c|}{\textbf{4.8}} & \multicolumn{1}{c|}{0.2} & 2.0 \\ \hline
toy truck$\ast$ & \multicolumn{1}{c|}{58.7} & \multicolumn{1}{c|}{57.5} & \textbf{76.5} & \multicolumn{1}{c|}{5.9} & \multicolumn{1}{c|}{0.0} & 2.8 & \multicolumn{1}{c|}{9.6} & \multicolumn{1}{c|}{0.0} & \textbf{11.3} & \multicolumn{1}{c|}{9.6} & \multicolumn{1}{c|}{0.0} & \textbf{11.3} & \multicolumn{1}{c|}{2.2} & \multicolumn{1}{c|}{0.1} & \textbf{4.2} & \multicolumn{1}{c|}{5.0} & \multicolumn{1}{c|}{0.2} & \textbf{12.9} & \multicolumn{1}{c|}{5.0} & \multicolumn{1}{c|}{0.2} & \textbf{12.9} \\ \hline
all & \multicolumn{1}{c|}{33.2} & \multicolumn{1}{c|}{52.7} & \textbf{87.8} & \multicolumn{1}{c|}{1.9} & \multicolumn{1}{c|}{4.5} & \textbf{15.8} & \multicolumn{1}{c|}{3.42} & \multicolumn{1}{c|}{9.2} & \textbf{32.4} & \multicolumn{1}{c|}{3.7} & \multicolumn{1}{c|}{9.3} & \textbf{32.6} & \multicolumn{1}{c|}{2.8} & \multicolumn{1}{c|}{0.3} & \textbf{26.8} & \multicolumn{1}{c|}{5.41} & \multicolumn{1}{c|}{0.9} & \textbf{43.8} & \multicolumn{1}{c|}{7.2} & \multicolumn{1}{c|}{0.9} & \textbf{44.4} \\ \hline
\end{tabular}

}
\vspace{-3mm}
\caption{Quantitative results on OO3D-9D. $\dagger$: continuous symmetric. $\ddagger$: discrete symmetric. $\ast$: non-symmetric.}
\label{tab:pca_ours}

\vspace{-5mm}
\end{table*}

\begin{figure*}[t]
\centering
\includegraphics[width=0.7\linewidth]{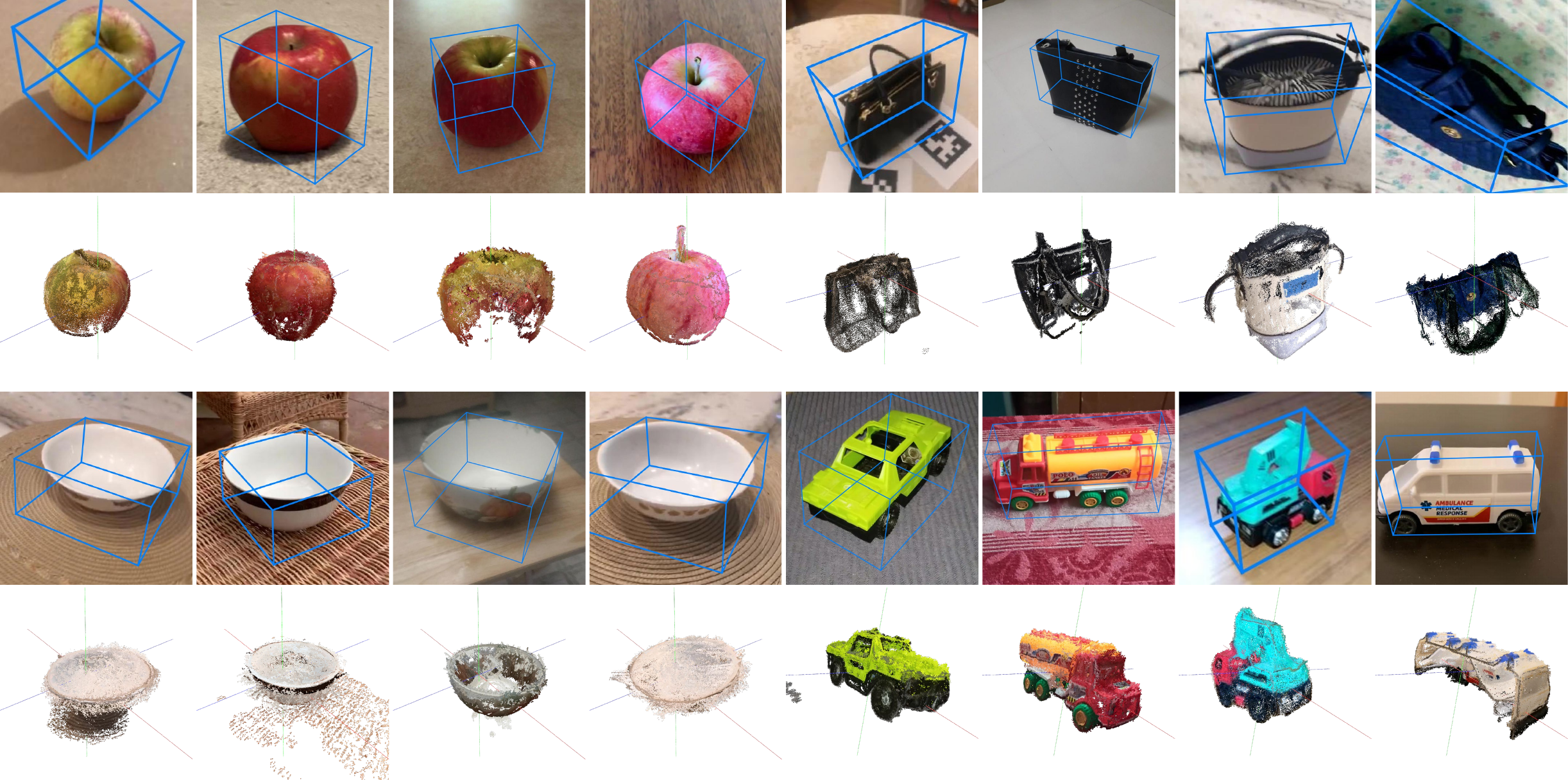}
\vspace{-12pt}
\caption{Qualitative results on unseen real-world data. Our model trained with OO3D-9D data could directly perform 6D pose and size estimation on the unseen Co3Dv2 dataset. The odd rows display cropped real-world images with our estimated 6D pose and size (visualized with the object bounding box), while the even rows display their corresponding aligned shapes in the normalized object coordinate space.}
\vspace{-3mm}
\label{fig:co3d}
\end{figure*}

\begin{figure*}
    \centering
    \includegraphics[width=0.8\linewidth]{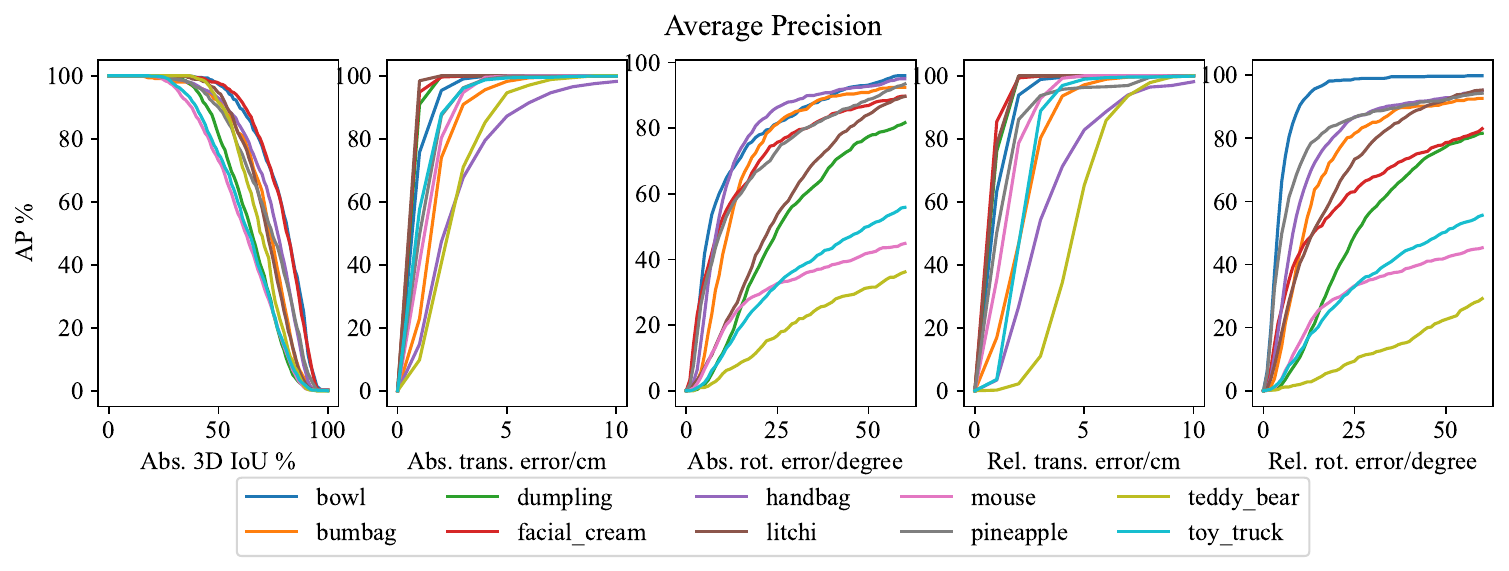}
    
    \vspace{-3mm}
    \caption{The absolute and relative precision on OO3D-9D. The average precision (AP) with respect to different thresholds of 3D IoU, rotation precision, and translation precision are reported. }
    \label{fig:ap_curves}
    \vspace{-3mm}
\end{figure*}

\section{Experiments}
\subsection{Implementation Details}

\textbf{Training details}. We train the model for 25 epochs with a batch size of 64 on 4 A100 GPUs. We use the AdamW~\cite{loshchilov2018decoupled} as the optimizer with a fixed learning rate $0.0001$. The input size of images is set to $480^2$. 
To compute the training loss in Eqn.~\ref{eqn:smooth_l1_loss_with_symmetry} for continuous symmetric objects, we approximate the augmented map set $\mathcal{S}$ by sampling $36$ angles around the rotation axis with equal angle interval. The sampled rotation matrices are applied to $S$, yielding an approximated map set $\mathcal{S}$. 
In practice, the input mask could be obtained by an open-vocabulary masking model~\cite{xu2023open} and the depth could be obtained by a depth estimation model~\cite{yuan2022neural}. In the experiments, for simplicity for evaluating our core modules, we use the ground-truth masks and depths provided in the dataset.

\textbf{Dataset rendering}. We employ the camera intrinsics used by the YCB-V~\cite{xiang2018posecnn} dataset as the parameters of the virtual camera in BlenderProc. The image resolution is set to $480 \times 640$, which is also in accordance with YCB-V.  

\subsection{Benchmark Datasets}

To thoroughly evaluate the performance of the models, we mainly conduct experiments on OO3D-9D single-object scenes with \textit{unseen categories}. Concretely,  we manually select 10 categories from the dataset as the test data, which are listed in Table~\ref{tab:pca_ours}. Overall, the dataset includes a total of 230 instances, comprising non-symmetric, discrete symmetric, and continuous symmetric objects. 

To demonstrate the effectiveness of our method, we further conduct experiments on Co3Dv2~\cite{reizenstein2021common} and visualize the prediction results of various novel instances. For further details on the data and experiments involving new viewpoints of known instances and new instances of known categories, please refer to the supplementary material.

\subsection{Evaluation Metrics}
\label{subsec:evalutation_metrics}
We first evaluate our results on the widely-used Rot\&Trans precision ($a^{\circ}\ b\ cm$)~\cite{li2018deepim}, and the 3D Intersection over Union (IoU)~\cite{geiger2013vision} metrics, and then evaluate with a new proposed metric, namely relative Rot\&Trans precision. 

\textbf{Rot\&Trans precision} is the average precision of the samples where the error of rotation is less than $a^{\circ}$ and the error of translation is less than $b\ cm$, as
\begin{equation}
\label{eqn:n_deg_m_cm}
\frac{1}{N}\sum_{j=1}^{N}{f_{ab}(\hat{{}^{c_j}T_{\mathcal{O}_i}}, {}^{c_j}T_{\mathcal{O}_i}, a, b)}, 
\end{equation}
where $f_{ab} \in \{0, 1\}$ is $1$ only if the pose differences are within the thresholds $a$ and $b$. 
We should note that for objects with discrete symmetry, we will first augment the ground truth poses according to the number of symmetric matrices, and then compare the estimated results with the augmented pose set. The pose with minimal error will be considered as the final ground truth pose. For objects with continuous symmetry, we only compare the angle difference between the ground truth and estimated principal axes for the computation of rotation error. 

\textbf{3D IoU} denotes the intersection over union (IoU) of two 3D bounding boxes. In this work, we follow~\cite{wang2019normalized} and compute axis-aligned 3D IoU@50 based on the Pytorch3D implementation~\cite{ravi2020pytorch3d}. 

\textbf{Relative Rot\&Trans precision.} 
Since the model may have its own cognition of the object reference frames, which might be different from the ones provided by the data, for the objects of unseen categories. We further propose a novel metric that can be used to evaluate the consistency of the estimated poses. Suppose the estimated poses for specific object category $\mathcal{O}_i$ are denoted as $({}^{c_1}T_{\mathcal{O}_ip_1}, {}^{c_2}T_{\mathcal{O}_ip_2}, ..., {}^{c_n}T_{\mathcal{O}_ip_n})$ and their corresponding ground truths are $({}^{c_1}T_{\mathcal{O}_i}, {}^{c_2}T_{\mathcal{O}_i}, ..., {}^{c_n}T_{\mathcal{O}_i})$. $\mathcal{O}_ip_j$ represents the predicted object reference frame that is recognized by the model for the $j$th sample. We can thus obtain the relative pose set from the predicted to the ground truth object frame, which is denoted as $({}^{\mathcal{O}_i}T_{\mathcal{O}_ip_1}, {}^{\mathcal{O}_i}T_{\mathcal{O}_ip_2}, ..., {}^{\mathcal{O}_i}T_{\mathcal{O}_ip_n})$. Then we compute the minimal relative difference among this pose set and regard it as the relative pose error. The average precision of the relative poses is then computed as
\begin{equation}
\label{eqn:relative_loss}
\frac{1}{N-1} \max_{j}\sum_{k=1,k \neq j}^{N}{f_{ab}({}^{\mathcal{O}_i}T_{\mathcal{O}_ip_k},{}^{\mathcal{O}_i}T_{\mathcal{O}_ip_j}, a, b)}.
\end{equation}

\subsection{Baselines}
\vspace{-3mm}
Since there is no work in the literature that researches this novel task, we find very few baselines that can be applied to this problem. The widely-used baselines such as ICP~\cite{besl1992method} or CPD~\cite{myronenko2010point} cannot be used in this task, since we assume that the 3D object model is not available during the inference stage. Since the principal component analysis (PCA) can infer the coordinate system according to the data observation, which is usually used as a baseline in the task of shape canonicalization~\cite{sajnani2022condor}, we consider it as the baseline to perform the estimation of translation and rotation from the observed point cloud. To evaluate the performance in comparison to other methods, we further apply the training data to IST-Net~\cite{Liu_2023_ICCV}, which is the current state-of-the-art method for close-set category-level object pose estimation, and assess the metrics on the test set with new categories.

\subsection{Evaluation Results}
\subsubsection{Results on OO3D-9D}
\label{subsubsec:results_on_oo3d_9d}
In this experiment, we sample 50 views for each object instance and evaluate the metrics for each category. The mean average precision (mAP) of all metrics is presented in Table~\ref{tab:pca_ours}, while Fig.~\ref{fig:ap_curves} shows more comprehensive results for each category.

Overall, our method significantly outperforms the baseline across all metrics. Specifically, our method attains mAPs of 87.8\%, 15.8\%, and 32.4\% for absolute IoU@50, $5^\circ5cm$, and $10^\circ5cm$, respectively, indicating the preference of the model for determining object reference frames aligned with manually labeled data.
On the other hand, we also obtain mAPs of 26.8\% and 43.8\% for relative $5^\circ5cm$ and $10^\circ5cm$, compared with 2.8\% and 5.4\% achieved by PCA, demonstrating that our method recognizes more consistent reference object frames within the same category. While IST-Net achieves slightly better results on absolute metrics than PCA, it performs worse on relative metrics, which shows that such close-set methods cannot directly adapt to unseen object categories even with large amounts of training data. 

Furthermore, we observe that our method attains mAPs of 44.0\%, 31.6\%, and 3.0\% for relative $5^\circ5cm$ on categories with continuous symmetric, discrete symmetric, and non-symmetric instances, respectively. We believe that symmetric objects tend to have regular shapes, making it easier for the model to capture their internal principal directions. 

\subsubsection{Results on Co3Dv2}
We visualize in Figure~\ref{fig:co3d} the qualitative results on unseen real-world data from Co3Dv2. The even rows display point clouds of instances transformed into normalized object coordinate space based on the estimation of our method. It can be observed that our open-vocabulary framework trained with only synthesized data can accurately estimate pose and size across instances with various novel categories. 
Interestingly, the upward directions of all instances align with the y-axes, consistent with most training data instances (some shown in Figure~\ref{fig:oo3d_9d_example}). This further suggests that by training with our OO3D-9D, the open-vocabulary framework can transfer the cognition of canonical object frames to novel objects in a manner consistent with human understanding.

\begin{table}[htb]
\centering
\small
\scalebox{0.8}{
\begin{tabular}{l|c|c|c|c}
\hline
 & REAL275 & 50 cat. & 100 cat. & All cat. \\ \hline
Abs IoU@50 & 48.2 & 75.5 & 81.2 & \textbf{87.8} \\ \hline
Abs $5^\circ5cm$ & 2.8 & 6.9 & 11.8 & \textbf{15.8} \\ \hline
Rel $5^\circ5cm$ & 4.5 & 7.8 & 21.9 & \textbf{26.8} \\ \hline
\end{tabular}}
\vspace{-3mm}
\caption{Ablation study of different scales of training data.}
\vspace{-2mm}
\label{tab:eff_oo3d9d}
\end{table}

\begin{table}[htb]
\centering
\small
\scalebox{0.8}{
\begin{tabular}{l|c|c}
\hline
 & Dino & Full (Dino+SD) \\ \hline
Abs IoU@50 & 75.0 & \textbf{87.8} \\ \hline
Abs $5^\circ5cm$ & 8.8 & \textbf{15.8} \\ \hline
Rel $5^\circ5cm$ & 9.8 & \textbf{26.8} \\ \hline
\end{tabular}}
\vspace{-3mm}
\caption{Framework ablation.}
\vspace{-3mm}
\label{tab:dino_sd}
\end{table}

\begin{table}[htb]
\centering
\small
\scalebox{0.8}{
\begin{tabular}{c|c|c|c}
\hline
 & Dummy & Free-form text & Category Labels \\ \hline
Abs $5^\circ5cm$ & 8.34 & 13.4 & \textbf{15.8} \\ \hline
Rel $5^\circ5cm$ & 11.2 & 21.9 & \textbf{26.8} \\ \hline
\end{tabular}}
\vspace{-3mm}
\caption{Ablation study of different text prompts}
\label{tab:prompt_impact}
\end{table}

\subsubsection{Ablation Study}
\label{subsubsec:ablation_study}

\noindent\textbf{Effects of different network components}. We first examine the effectiveness of different encoder modules of the proposed framework. Specifically, we train the model using only the Dino encoder and compare it to the full framework (Dino + Stable Diffusion), yielding the results in Table~\ref{tab:dino_sd}. It is noticeable that while the Dino module performs well on various visual tasks such as depth estimation~\cite{oquab2023dinov2}, directly adapting the features for this task results in inferior performance, which only achieves mAPs of 8.8\% and 9.8\% on the absolute and relative $5^\circ5cm$ precision, respectively.

\noindent\textbf{Effects of the scale of the training data}. We ablate on the number of object categories used by our method to assess the impact of the data scale. We randomly select 2 subsets from the original OO3D-9D dataset, containing 50 and 100 object categories, respectively. We then train our model on these two sub-datasets and evaluate the performance of data with novel categories. We also train our method on the NOCS REAL275 dataset and perform evaluation on the same test data. 
 The results, presented in Table~\ref{tab:eff_oo3d9d}, indicate that performance improves as the number of training categories increases, demonstrating that a larger dataset can enhance the transfer ability of open-vocabulary category level pose estimation.

\begin{figure}[h]
\centering
\includegraphics[width=1.0\linewidth]{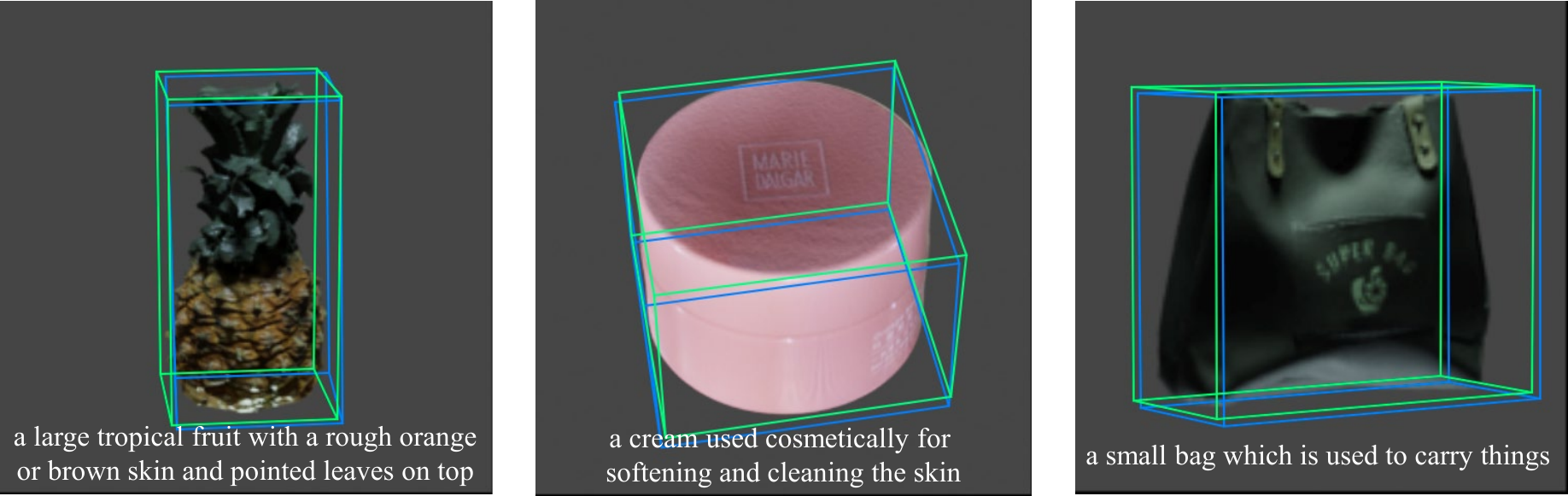}
\vspace{-5mm}
\caption{Visualization of results with free-form text prompts.}
\label{fig:free_form_text}
\vspace{-3mm}
\end{figure}

\noindent\textbf{Effects of text descriptions}. We investigate the influence of prompts by utilizing free-form descriptions to characterize objects. We also use dummy prompts with meaningless spaces for comparison. Specifically, we substitute object names with descriptions sourced from Wikipedia and employ the textual features of these sentences to guide the estimation of NOCS maps for the target objects. Some of the text descriptions and their corresponding estimated results are depicted in Fig.~\ref{fig:free_form_text} and the results are shown in Table~\ref{tab:prompt_impact}. Thanks to the prior knowledge from Diffusion UNet which is pre-trained with large-scale diverse image-text pairs, our method still gains effective information from these free-form texts. 

\section{Conclusions and Limitations}
This paper presents a new open-set problem for estimating object pose and size from RGBD images and text descriptions. We propose an open-vocabulary framework that predicts NOCS maps using foundation models and develop OO3D-9D, a large-scale, photorealistic dataset with diverse categories.
Experiments show that the proposed method trained with OO3D-9D can be effectively generalized to unseen object categories. However, our method achieves inferior performance when faced with non-symmetric object instances or categories with large intra-class variance. Moreover, this method relies on precise object masks to crop out the region of the target objects and the scenes in the training data are relatively simple compared with multi-object scenes. Future work will consider leveraging this dataset and integrating object detection and pose estimation in a unified framework.  


\appendix
\section{Appendix}

\subsection{Legitimacy of evaluation metrics}
\begin{figure}[h]
    \includegraphics[width=\linewidth]{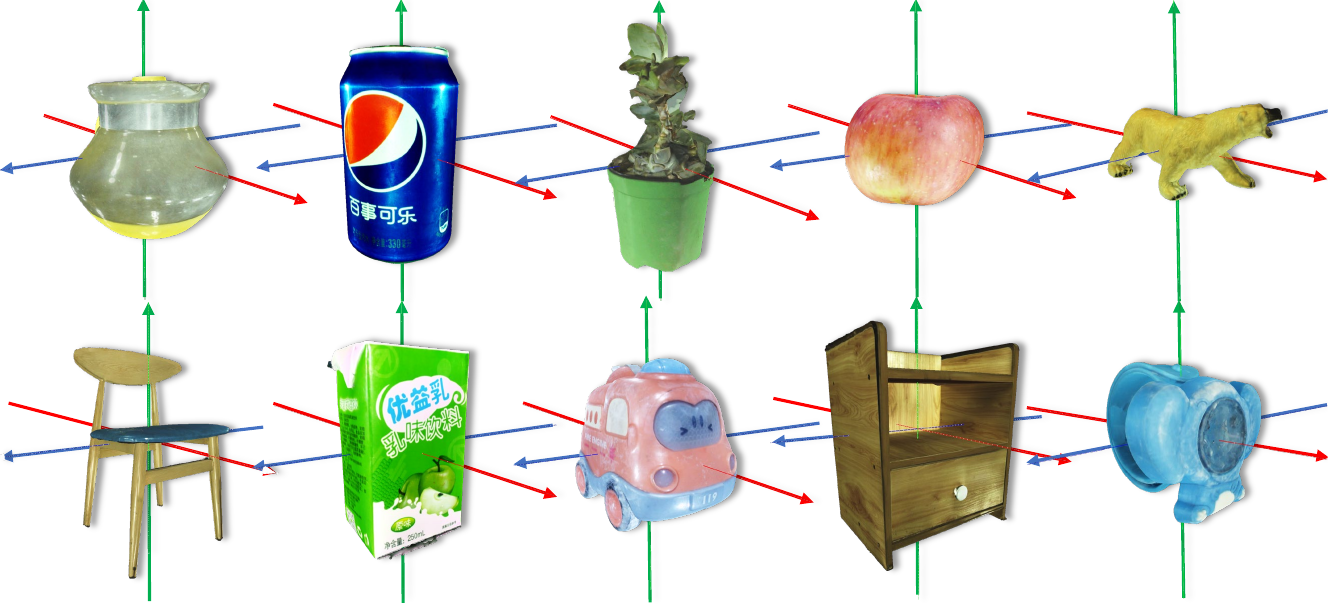}
    \vspace{-3mm}
    \caption{Visualization of instances in OmniObject3D with coordinate frames. In these examples, it is obvious to see that the upward directions are aligned with y-axes (green arrows) and the front directions are consistent with the x-axes (red arrows).}
    \label{fig:ref_frames}
    \vspace{-3mm}
\end{figure}

\begin{figure}[h]
    \includegraphics[width=\linewidth]{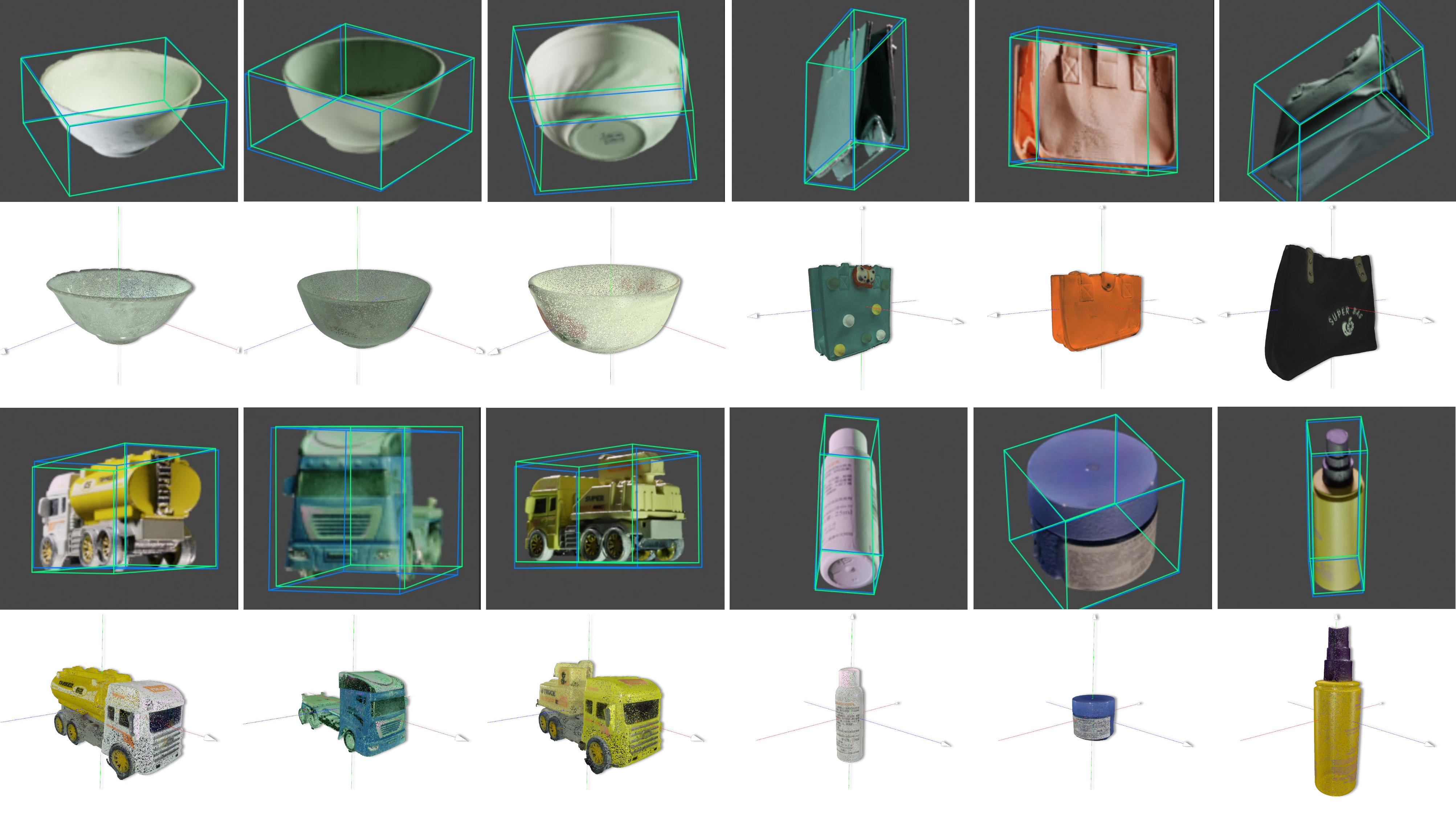}
    \vspace{-3mm}
    \caption{Qualitative results on unseen data. The odd rows display the estimated (blue boxes) and the ground-truth (green) poses, while the even rows display their estimated aligned shapes in the normalized object coordinate space.}
    \label{fig:unseen_oo3d9d}
    \vspace{-3mm}
\end{figure}

It is worth noting that in the setting of open-vocabulary category-level object pose estimation problem, the model should have its own cognitive reference frame for each category, and these reference frames might be different from the ones manually annotated by human beings, which makes the absolute metrics mentioned in Sec.~4.3 inappropriate for this task. 
However, in OmniObject3D, the reference frames of most objects are aligned with specific patterns, e.g., the upward directions of the instances in OmniObject3D are usually aligned with the y-axes, and the front directions are usually aligned with the x-axes. 
Example object instances with labeled reference frames in these patterns are visualized in Figure~\ref{fig:ref_frames}. 
These patterns could be learned by the model in the training, and generalized to unseen categories, as visualized in Figure~\ref{fig:unseen_oo3d9d}. 
Therefore, the absolute metrics are still a good indicator for evaluating the model in learning the patterned poses, which also benefits the subsequent applications like grasping.

\subsection{Applications}
In this section, we will showcase potential applications stemming from this task, highlighting the significance of conducting research on it. Our focus will primarily be on two tasks: robotic grasping and object reconstruction. It is essential to reiterate that, due to the formulation of open-vocabulary category-level object pose estimation, the canonical frames of instances within the same category must be consistent. Furthermore, instances from different categories with similar attributes will possess analogous canonical frames, resulting from the alignment learned through the language model and manually labeled data.

\begin{figure}[h]
    \includegraphics[width=\linewidth]{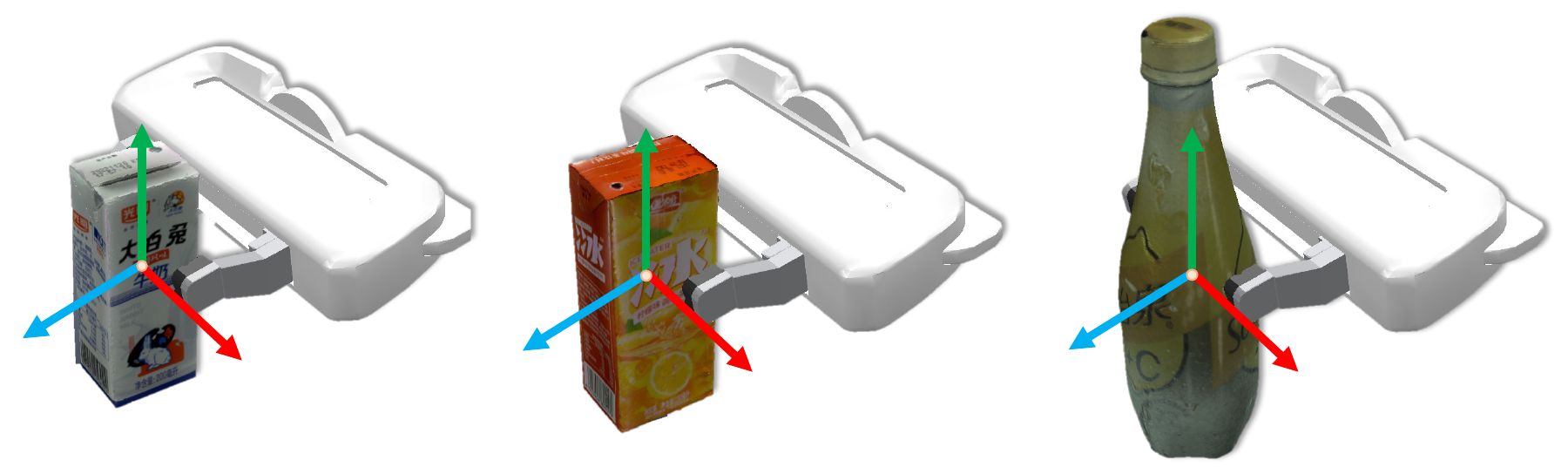}
    \vspace{-3mm}
    \caption{Illustration of cross-instance / cross-category grasping. Left: pre-defined gripper pose for the box beverage instance. Middle: transferred gripper pose for the new box beverage instance. Right: transferred gripper pose for the bottle instance. }
    \label{fig:grasping}
    \vspace{-3mm}
\end{figure}

\noindent{\textbf{Robotic grasping}}. 
Building upon the aforementioned setup, we can transfer grasping knowledge to novel instances. An example is provided in Figure~\ref{fig:grasping}. Given an object instance from a specific category with predefined grasp poses (left), we can directly transfer the gripper poses to new instances within the same category (middle) once the object pose of this instance is recognized. Likewise, we can also transfer them to unseen instances (right) whose object attributes resemble the predefined instance.

\begin{figure}[h]
    \includegraphics[width=\linewidth]{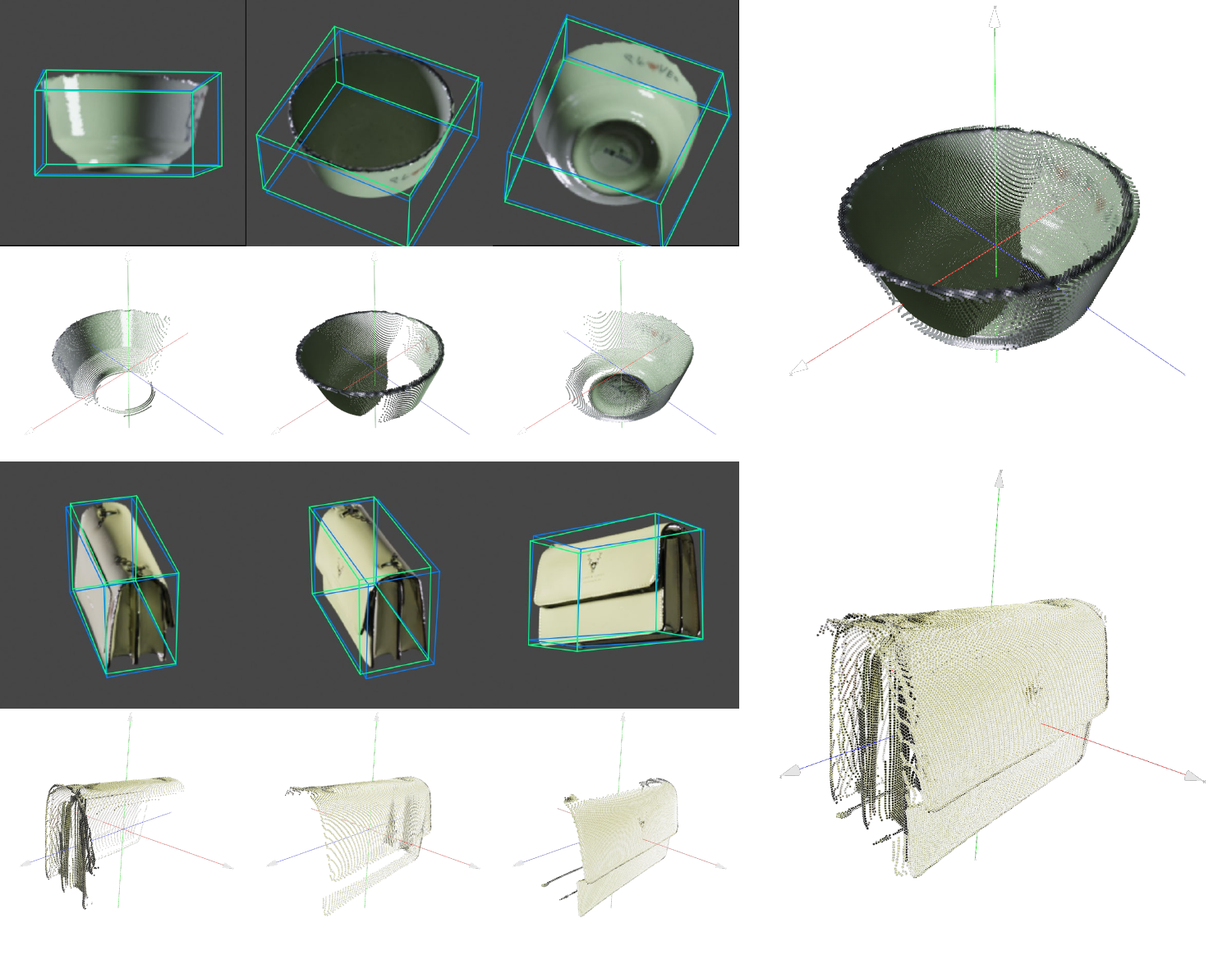}
    \vspace{-3mm}
    \caption{Examples of object reconstruction. }
    \label{fig:reconstruction}
    \vspace{-3mm}
\end{figure}

\noindent{\textbf{Object reconstruction}}. 
As the model possesses a unique reference frame for a specific object instance, we can utilize the recognized poses from the model to perform object reconstruction. Figure~\ref{fig:reconstruction} displays the results of combined point clouds for the same instances, where the partial point clouds originate from depth maps of various views and are aligned by the estimated object poses.

\subsection{Visualization of the data}
\begin{figure*}[t]
     \centering
     \begin{subfigure}[b]{0.33\textwidth}
         \centering
         \includegraphics[width=\textwidth]{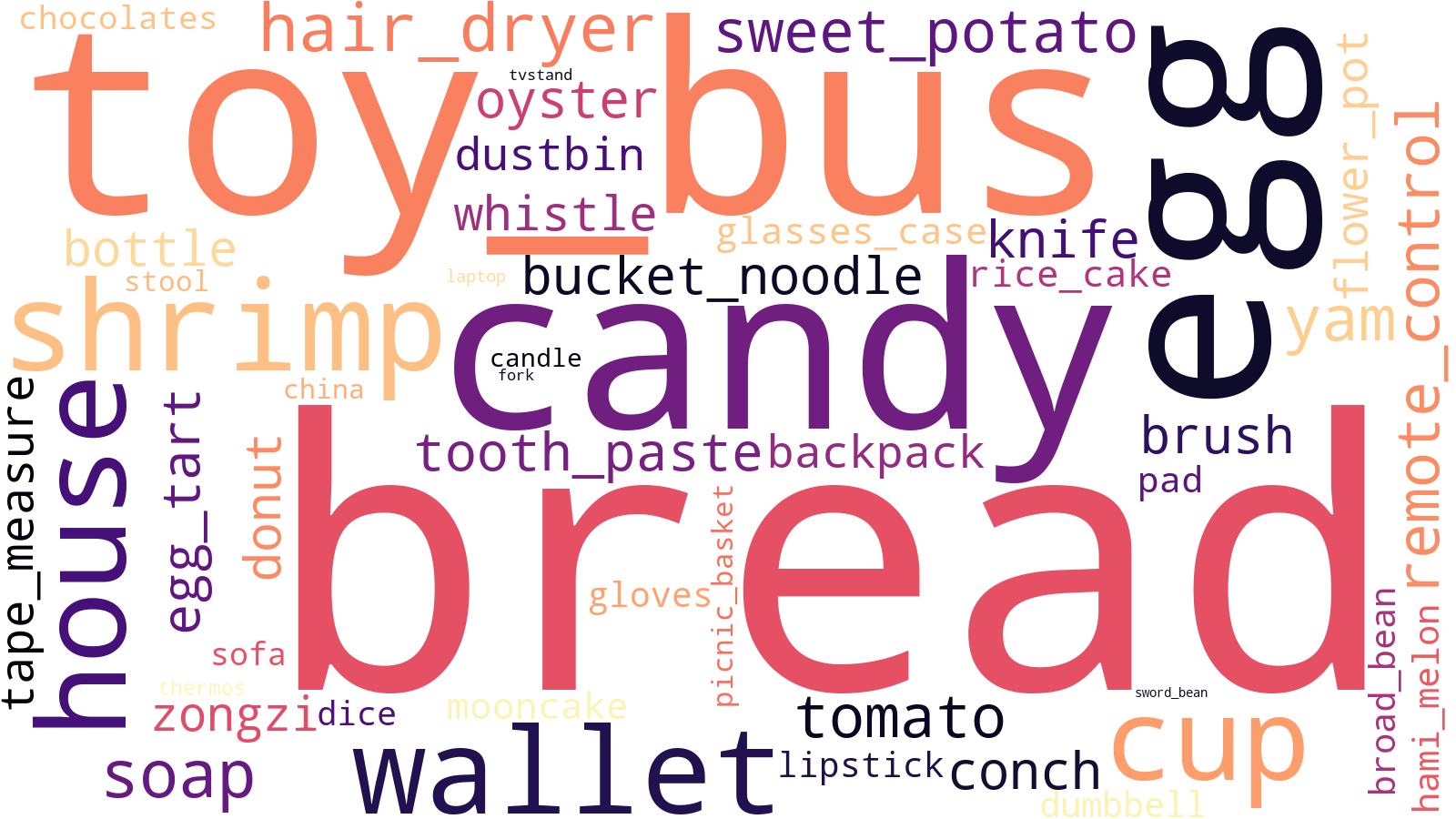}
         \caption{Set1: 50 object categories}
         \label{fig:wc50}
     \end{subfigure}
     \begin{subfigure}[b]{0.33\textwidth}
         \centering
         \includegraphics[width=\textwidth]{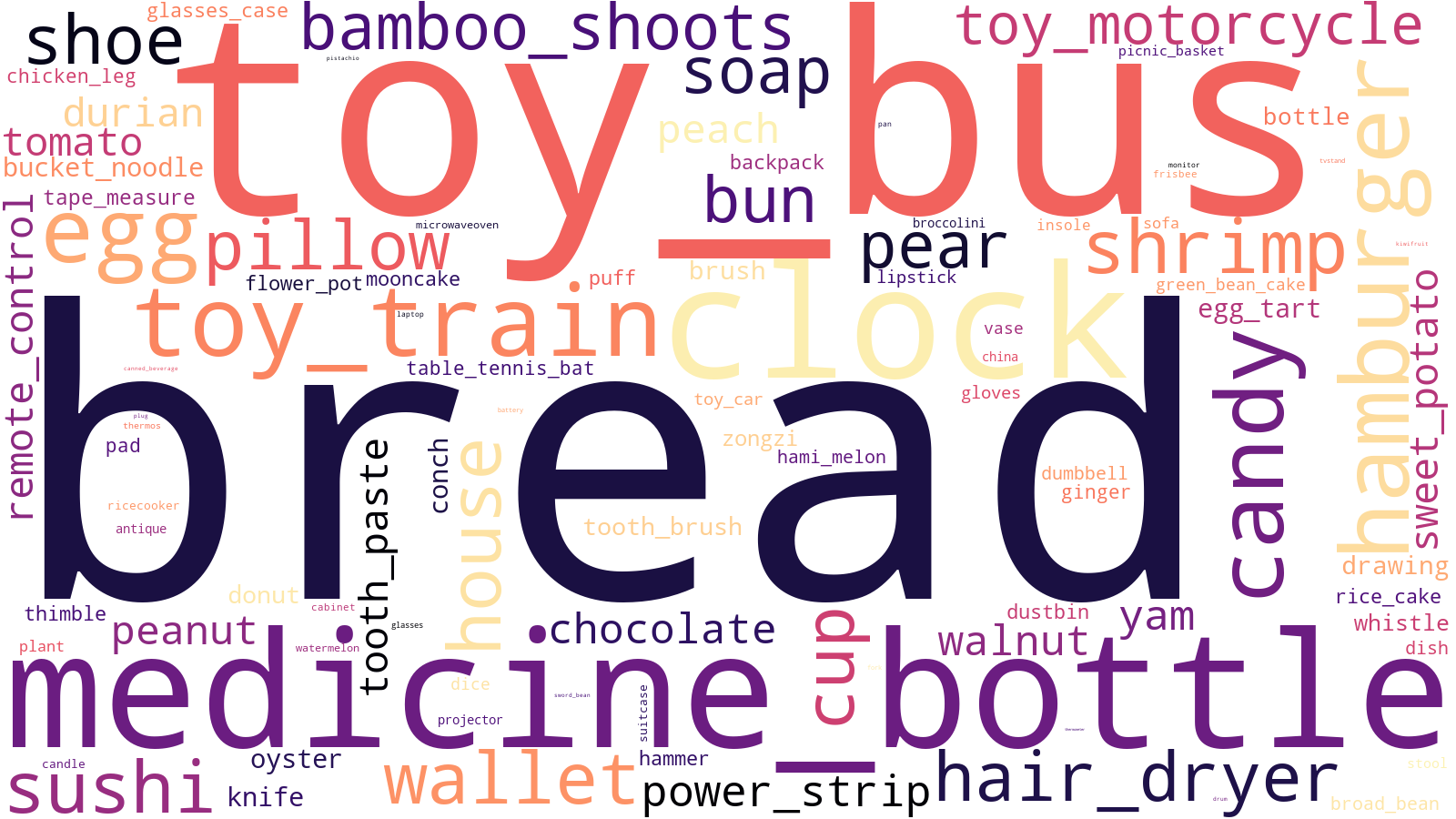}
         \caption{Set2: 100 object categories}
         \label{fig:wc100}
     \end{subfigure}
     \begin{subfigure}[b]{0.33\textwidth}
         \centering
         \includegraphics[width=\textwidth]{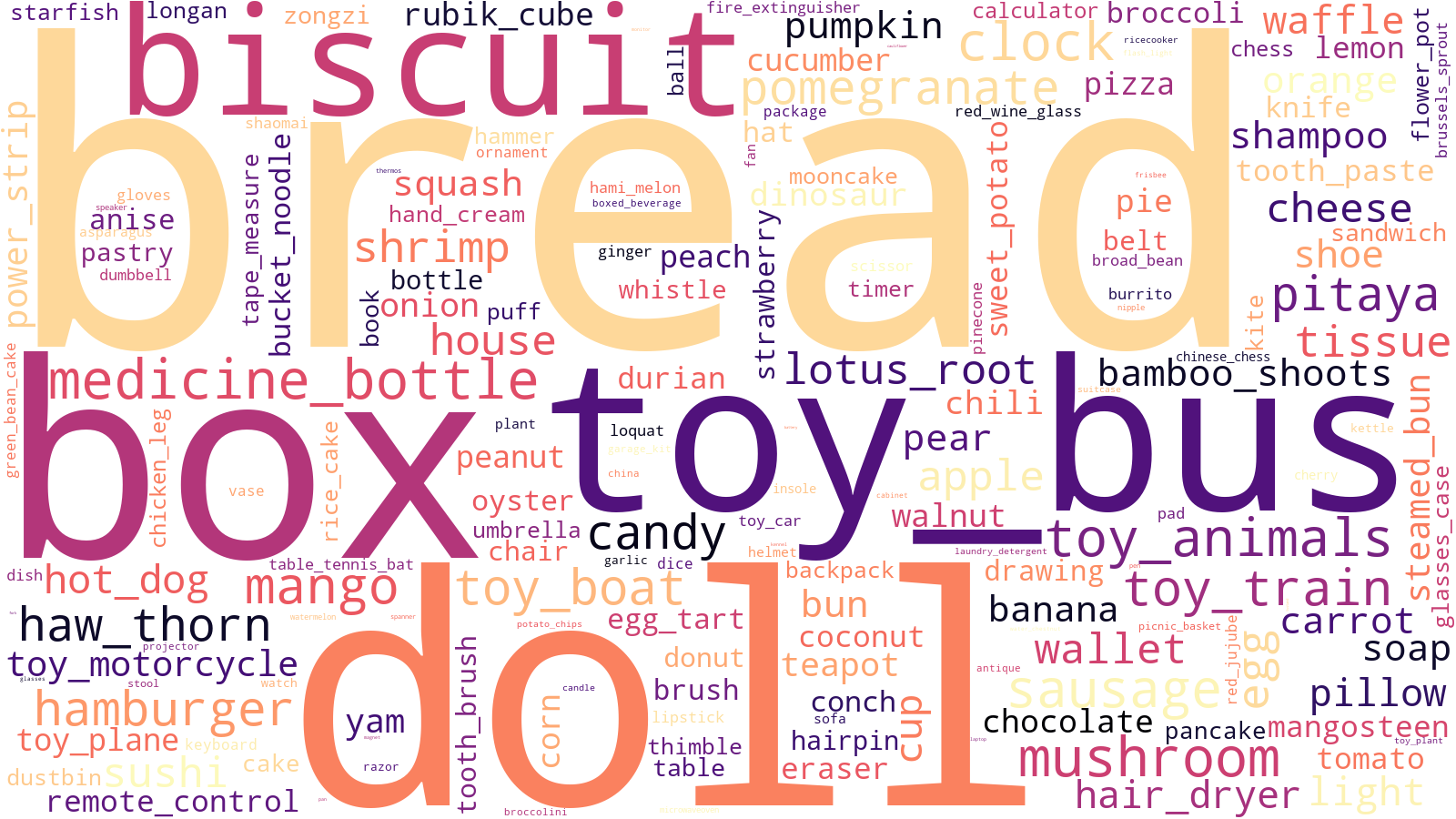}
         \caption{Set3: 206 object categories}
         \label{fig:wc204}
     \end{subfigure}
    \caption{Visualization of the object categories in the OmniObject3D dataset. Larger font size indicates more examples of that object category.}
    \label{fig:oo3d_cats}
\end{figure*}

In Sec.~4.5.3, we study the effect of the data scale on the performance. Figure~\ref{fig:oo3d_cats} visualizes the specific categories we used when performing ablation analysis about data scale. The number of instances including the number of non-symmetric, discrete symmetric, and continuous symmetric instances for each subset are listed in Table~\ref{tab:data_statistics}.  

\begin{figure}[h]
    \includegraphics[width=\linewidth]{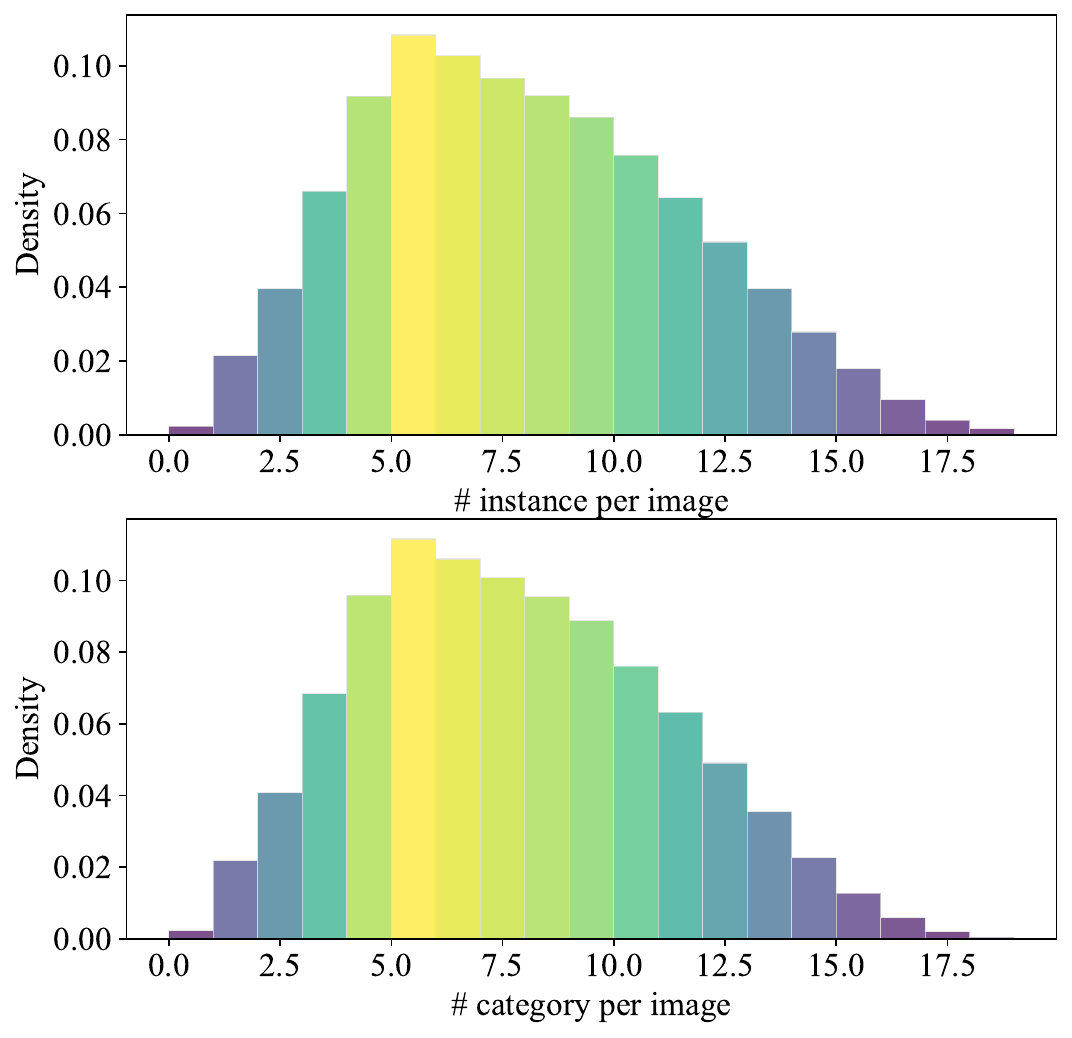}
    \vspace{-3mm}
    \caption{Statistics of the multi-object dataset. }
    \label{fig:statistics_oo3d_multi}
    \vspace{-3mm}
\end{figure}

For the multi-object dataset, we also count the distributions of the number of instances and the number of object categories per image among 50k samples. The density distributions are visualized in Figure~\ref{fig:statistics_oo3d_multi}. 

\begin{table}[h]
\centering
\begin{tabular}{c|c|c|c|c|c}
\hline
      & \# cat. & \# ins. & \# NS & \# DS & \# CS \\ \hline
Set 1 & 50      & 1290    & 649   & 348   & 331   \\ \hline
Set 2 & 100     & 2291    & 1127  & 513   & 689   \\ \hline
Set 3 & 204     & 4782    & 1982  & 1004  & 1904  \\ \hline
\end{tabular}

\footnotesize{NS: non-symmetric. DS: discrete symmetric. CS: continuous symmetric.}
\vspace{-3mm}
\caption{Statistics of data.}
\label{tab:data_statistics}
\end{table}

\begin{table}[h]
\centering
\scalebox{0.7}{
\begin{tabular}{c|c|c|c|c|c}
\hline
 &  & Abs IoU@50 & Abs $5^\circ5cm$ & Abs $10^\circ5cm$ & Abs $10^\circ10cm$ \\ \hline
\multirow{4}{*}{\begin{tabular}[c]{@{}c@{}}New \\ views\end{tabular}} & NS & 97.1 & 37.6 & 65.9 & 66.5 \\ \cline{2-6} 
 & DS & 97.5 & 46.7 & 73.9 & 74.4 \\ \cline{2-6} 
 & CS & 99.0 & 59.2 & 82.2 & 82.3 \\ \cline{2-6} 
 & All & 97.9 & 47.6 & 73.7 & 74.0 \\ \hline
\multirow{4}{*}{\begin{tabular}[c]{@{}c@{}}New \\ instances\end{tabular}} & NS & 93.0 & 23.5 & 46.3 & 47.2 \\ \cline{2-6} 
 & DS & 96.1 & 44.0 & 68.6 & 68.7 \\ \cline{2-6} 
 & CS & 97.3 & 45.1 & 65.4 & 65.4 \\ \cline{2-6} 
 & All & 95.5 & 36.7 & 58.8 & 59.2 \\ \hline
\end{tabular}}
\vspace{-3mm}
\caption{Extra results of OO3D-9D.}
\label{tab:oo3d9d_uv_ko_uo_kc}
\end{table}

\subsection{More results on OO3D-9D}
We conduct extra experiments on OO3D-9D by evaluating the metrics on known objects with novel views and unseen objects with known categories. To test the metrics on known objects with novel views, we randomly sample 50 novel views for each instance and perform the evaluation on these new samples. To construct the new instance set, we randomly select 2 instances from an object set of a specific category if the number of instances is greater than 20. The number of test views is 50 as well. The results are shown in Table~\ref{tab:oo3d9d_uv_ko_uo_kc}. It is obvious to see that 1) the model performs better on objects with known categories, and it achieves the best performance when dealing with known instances, 2) similar to the results in Sec.~4.5.1, our method achieves better precision on objects with symmetry for both test sets, which further implies the challenge of non-symmetric objects for the task of open-vocabulary category-level object pose estimation. 

To evaluate the close-set performance in comparison to other methods, we apply the training data to IST-Net~\cite{Liu_2023_ICCV}, which is the current state-of-the-art method for close-set category-level object pose estimation, and assess the metrics on the test set with new instances. Specifically, we maintain a training batch size of 64, keeping other settings unchanged. The results can be found in Table~\ref{tab:istnet}. It is obvious to see that the performance of IST-Net is much inferior to our method, as it employs ResNet18~\cite{he2016deep} and PointNet++~\cite{qi2017pointnet++} as the network backbones, which are capability-limited when trying to fit the pattern from our dataset containing diverse object categories. 

\begin{table}[h]
\centering
\scalebox{0.7}{
\begin{tabular}{c|c|c|c|c}
\hline
 & Abs IoU@50 & Abs $5^\circ5cm$ & Abs $10^\circ5cm$ & Abs $10^\circ10cm$ \\ \hline
IST-Net\cite{Liu_2023_ICCV} & 56.6 & 8.3 & 15.1 & 15.1 \\ \hline
Ours & \textbf{95.5} & \textbf{36.7} & \textbf{58.8} & \textbf{59.2} \\ \hline
\end{tabular}}
\caption{Comparison with IST-Net on OO3D-9D test set with new instances}
\label{tab:istnet}
\end{table}

\subsection{Implementation Details}
\subsubsection{Details of generating text features}
In this work, we borrow the prompt engineering used by~\cite{Zhao_2023_ICCV} to generate text features for each object category. Concretely, we substitute each category name into the ImageNet template containing 80 sentences. Each sentence is then fed into the CLIP textual encoder to generate the corresponding feature. We compute the mean of the features within the same category and consider it as the final feature for that category. For the experiments with free-form texts, we use the descriptions looked up from the Collins English Dictionary to replace the original object names. Examples of the descriptions are listed in Table~\ref{tab:free_form_text}. 

\begin{table*}[]
\centering
\begin{tabular}{l|l}
\hline
Object Name & Description                                    \\ \hline
bowl        & a round container                              \\ \hline
bumbag      & a small bag worn on a belt, round the waist    \\ \hline
dumpling    & a small lump of dough that is cooked and eaten \\ \hline
facial\_cream & a cream used cosmetically for softening and cleaning the skin                                      \\ \hline
handbag       & a small bag which a woman uses to carry things such as her money and keys in when she goes out     \\ \hline
litchi        & a small rounded fruit with sweet white scented flesh, a large central stone, and a thin rough skin \\ \hline
mouse       & a device that is connected to a computer       \\ \hline
pineapple     & a large tropical fruit with a rough orange or brown skin and pointed leaves on top                 \\ \hline
teddy\_bear   & a toy of children, made from soft or furry material, which looks like a friendly bear              \\ \hline
toy\_truck    & a mini version of a real truck, designed for play and fun                                          \\ \hline
\end{tabular}
\vspace{-3mm}
\caption{Examples of free-form text description.}
\label{tab:free_form_text}
\end{table*}

{
    \small
    \bibliographystyle{ieeenat_fullname}
    \bibliography{main}
}

\end{document}